%% file: look_closer.tex
\documentclass[10pt,twocolumn,a4paper]{article}
\usepackage[T1]{fontenc}
\usepackage[utf8]{inputenc}
\usepackage[margin=0.72in,columnsep=0.25in]{geometry}
\usepackage{mathptmx}
\usepackage{courier}
\usepackage{amsmath,amssymb}
\usepackage{graphicx,booktabs,array,tabularx}
\usepackage[table]{xcolor}
\usepackage{enumitem}
\usepackage{microtype}
\usepackage[numbers,sort&compress]{natbib}
\usepackage[font=small,labelfont=bf,hypcap=false]{caption}
\usepackage{fancyhdr}
\usepackage{pdflscape}
\usepackage{xurl}
\usepackage[colorlinks=true,linkcolor=blue!45!black,citecolor=blue!45!black,urlcolor=blue!45!black]{hyperref}
\hypersetup{pdftitle={Look Closer: Patch-wise Supervision for AI-Generated Image Detection},
 pdfauthor={Zhida Zhang, Tao Wu, Siyu Liu, Jie Cao}}
\setlist[itemize]{leftmargin=*,topsep=3pt,itemsep=1pt,parsep=0pt}
\setlist[enumerate]{leftmargin=*,topsep=3pt,itemsep=1pt,parsep=0pt}

\title{\vspace{-1.0em}\LARGE\bfseries Look Closer: Patch-wise Supervision\\for AI-Generated Image Detection}
\author{
Zhida Zhang$^{1}$ \quad Tao Wu$^{1,2}$ \quad Siyu Liu$^{3}$ \quad Jie Cao$^{1,*}$\\[5pt]
\small $^1$MAIS \& NLPR, Institute of Automation, Chinese Academy of Sciences (CASIA), Beijing, China\\
\small $^2$ShanghaiTech University, Shanghai, China \qquad
$^3$Anhui University, Hefei, China\\[4pt]
\small \texttt{zhida.zhang@cripac.ia.ac.cn}, \texttt{wutao2022@shanghaitech.edu.cn}\\
\small \texttt{liusiyu0102@gmail.com}, \texttt{jie.cao@cripac.ia.ac.cn}\\
\small $^*$Corresponding author
}
\date{}
\begin{document}
\raggedbottom
\maketitle
\thispagestyle{plain}
\begin{abstract}
How much of an image does a detector need to see? Small RGB regions can retain useful evidence of image synthesis even when they reveal little of the full scene. Motivated by single-patch detection, we study patch-wise supervision: a shared backbone classifies explicit crops, each crop receives its own loss, and patch probabilities are averaged only at inference. The procedure requires neither handcrafted residual filtering nor a learned image-level fusion module. Experiments span single-patch selection, multiple generator collections, and four CNN and Transformer backbones. On GenImage, the reported means are higher for the patch-wise variants than for their whole-image counterparts across all four backbones. Comparisons of supervision granularity, source resolution, crop size, and inference coverage further characterize the approach, while post-processing tests and difficult-image evaluation reveal its limitations. Some checkpoints were selected using target evaluation data, limiting conclusions about unbiased cross-generator generalization. Overall, the study identifies explicit local input and patch-level supervision as a simple, useful combination for investigating generalizable AI-generated image detection.
\end{abstract}
\begin{center}\small Code: \url{https://github.com/LF-Jade/look-closer}\end{center}

\input{sections/introduction}
\input{sections/method}
\input{sections/experiments}
\input{sections/discussion}
\begingroup
\small
\setlength{\bibsep}{3pt}
\bibliographystyle{plainnat}
\bibliography{references}
\endgroup
\clearpage
\onecolumn
\appendix
\input{sections/implementation}
\input{sections/appendix_tables}

\clearpage
\input{sections/exploratory}
\end{document}

%% file: sections/introduction.tex
\section{Introduction}
\label{sec:intro}

\begin{figure*}[t]
\centering
\includegraphics[width=\textwidth]{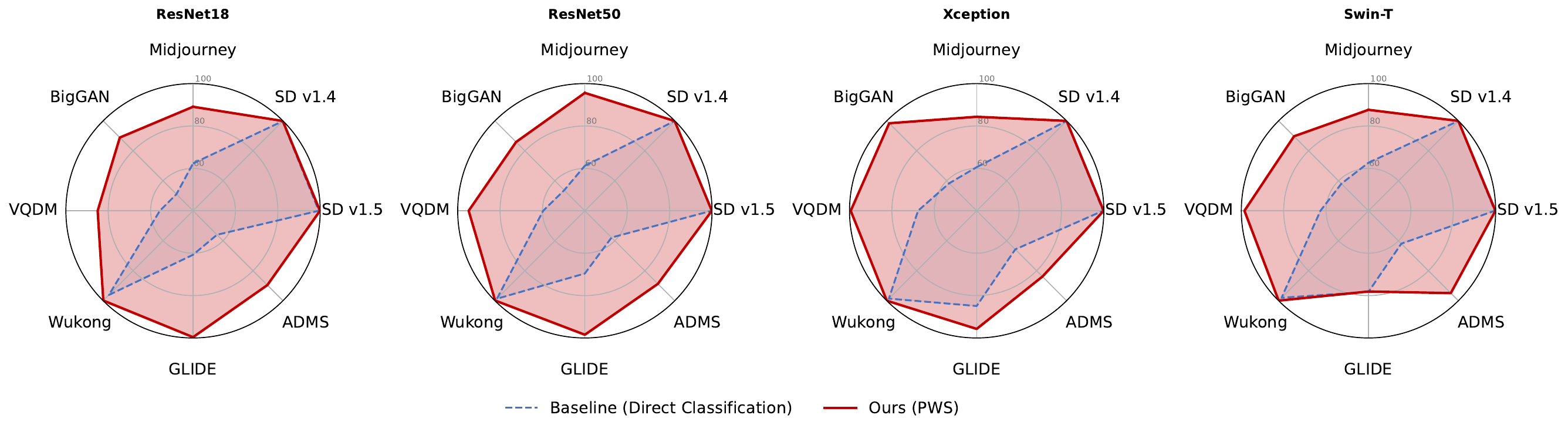}
\caption{GenImage results across four backbones. Blue dashed curves denote whole-image baselines; red curves denote PWS. The reported PWS mean is higher for each backbone, although this ordering does not hold for every generator subset. Exact values and the model-selection qualifications are given in Table~\ref{tab:GenImage} and Section~\ref{sec:selection}.}
\label{fig:teaser-radar}
\end{figure*}

A visually convincing image need not be locally indistinguishable from a photograph. Image synthesis can reproduce recognizable objects and coherent scenes while leaving differences in texture, reconstruction statistics, or relationships between neighboring pixels. For AI-generated image detection, this raises a practical question: how much of an image should a detector see at once? A whole-image classifier is a natural default, but its input combines local appearance with object identity and scene composition. Resizing that input may also alter the fine-scale evidence that detection relies on.

We study a deliberately simple alternative: learn from small RGB patches, and combine their predictions only when making the final image-level decision. This choice restricts the spatial context of each prediction without specifying a handcrafted forensic representation. It also changes the unit of supervision. Rather than asking a collection of patches to produce one correct training prediction, we ask each sampled patch to predict its parent image's label.

The study grew out of our earlier EIB-Net~\citep{eib2026}, which selects a low-complexity patch and regularizes its representation with an information bottleneck. A detector based on one selected region raises two further questions. Is useful evidence confined to that selection, or can other regions also be informative? If several regions contain useful but uneven evidence, can a simple multi-patch procedure exploit them without an additional fusion network?

Our single-patch experiments motivate this extension. In a facial-image setting, randomly selected patches perform similarly to minimum-complexity patches, while maximum-complexity patches also retain considerable detection ability. This does not mean that all regions are equally useful. Rather, committing to one selection heuristic is not the only way to obtain informative inputs. Multiple crops offer broader spatial coverage; separate patch losses encourage the classifier to use each local observation.

We study \emph{patch-wise supervision} (PWS): a shared classifier processes explicit RGB crops, each crop receives its own classification loss, and crop probabilities are averaged at inference. There is no learned within-image feature fusion, no requirement for a residual filter, and no need to truncate the backbone. The defining choice is to apply the loss to each patch before any image-level aggregation.

The experiments examine this recipe across ResNet-18, ResNet-50, Xception, and Swin-T on GenImage and AIGCDetectBenchmark. Supervision and resolution-alignment studies distinguish extracting crops from how those crops are trained. Patch-size and patch-count experiments probe the trade-off between local detail and coverage. Image-processing tests and Chameleon expose important failure modes. We report these existing results with their model-selection limitations, rather than present a newly standardized leaderboard evaluation.

Our contributions are:
\begin{itemize}
\item \textbf{From one region to multiple local decisions.} We connect single-patch observations to a multi-patch detector using RGB inputs, a shared full backbone, and inference-time probability averaging.
\item \textbf{Supervision as a design choice.} We formulate and compare patch-level supervision with an image-level objective applied after pooling, separating this choice from patch extraction.
\item \textbf{Evidence across architectures and conditions.} We study backbone choice, resolution, crop size, coverage, and post-processing, with detailed results and failure cases that make the scope of the approach visible.
\end{itemize}

\section{Related Work}
\label{sec:related}

\paragraph{Signals for generalizable detection.}
CNN detectors can transfer beyond their training generator under suitable data processing~\citep{wang2020}. Other approaches exploit pretrained representation spaces~\citep{ojha2023}, diffusion reconstruction error~\citep{dire}, or neighboring-pixel relationships associated with upsampling~\citep{npr}. These methods differ in how detection evidence is represented. PWS leaves the representation to a conventional RGB classifier and instead constrains spatial input and supervision.

\paragraph{Local prediction and patch selection.}
Patch Forensics~\citep{chai2020} restricts CNN receptive fields, applies local classification losses, and averages probabilities at inference. Its experiments include fully generated and manipulated facial images. We impose locality through explicit crops while retaining the backbone's feature-extraction stages, and examine natural-image generator collections and a windowed Transformer alongside CNNs. This is an implementation and empirical-scope distinction, not a first claim for local supervision.

PatchCraft~\citep{patchcraft} reorganizes texture patches and exploits rich- and poor-texture regions. SSP~\citep{ssp2024} selects a low-complexity patch and uses SRM filtering; its ablations also include direct RGB-patch classification without SRM. Thus, explicit cropping and RGB-patch classification have precedents. Panoptic Patch Learning~\citep{ppl2026} studies distributed evidence through patch interventions and a contrastive objective. PWS focuses on a minimal alternative: multiple separate RGB classifications, each supervised before within-image aggregation.

\paragraph{Connection to EIB-Net.}
EIB-Net~\citep{eib2026} is our preceding work on entropy-guided single-patch selection with a variational information bottleneck. Here we move from selecting one region toward learning from several regions with the same classifier. The present single-patch settings, previously denoted SPD, are not the complete EIB-Net system; the multi-patch method does not require the bottleneck. We retain their respective results rather than reinterpret cross-paper score differences as a controlled ablation of patch count.

\paragraph{Evaluation settings.}
GenImage~\citep{genimage} and AIGCDetectBenchmark~\citep{patchcraft} evaluate detection across generator families. DIFF~\citep{diff} and DiffusionForensics~\citep{dire} provide supporting facial-image and cross-content experiments. Chameleon~\citep{aide} emphasizes a more challenging distribution of high-quality images. Together, these settings reveal both encouraging benchmark behavior and cases where a local-input detector remains unreliable.

%% file: sections/method.tex
\section{From Single-Patch Evidence to Multiple Observations}
\label{sec:motivation}

\subsection{Local input as an inductive bias}
\begin{figure}[t]
\centering
\includegraphics[width=\linewidth]{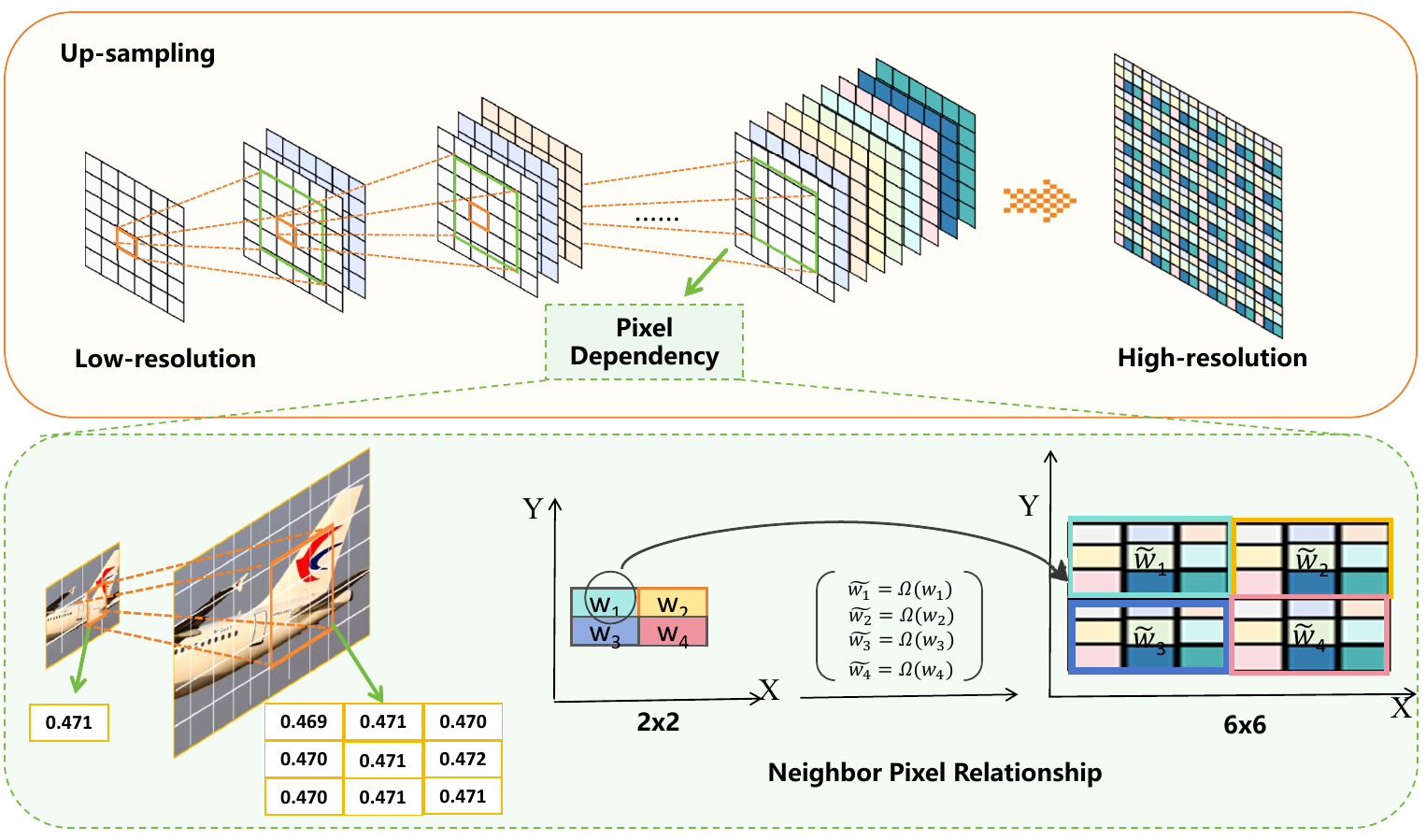}
\caption{Motivation for examining local pixel relationships. Upsampling can map a low-resolution representation into structured neighborhoods. The illustrated mapping and pixel values are schematic, not measurements establishing a universal artifact or a model of all generators.}
\label{fig:local-motivation}
\end{figure}

Generation and reconstruction operations can induce structured relationships between nearby output pixels. Figure~\ref{fig:local-motivation} illustrates this intuition for upsampling-based generation, following neighboring-pixel analysis~\citep{npr}. Such relationships motivate examining small regions even when an image is globally plausible. They do not imply a universal signature: photographs also have spatial dependencies, and both classes are affected by camera processing, compression, and resizing.

Our working hypothesis is that limiting spatial context can make local evidence easier to use while reducing access to full-image composition. This is an input constraint, not a guarantee of semantic-free features. The gradient co-occurrence analysis in Appendix~\ref{app:exploratory} offers a complementary view of local image statistics; it is not a component of the detector.

\subsection{What the single-patch study tests}
The documented single-patch protocol uses a $256\times256$ image canvas and a $4\times4$ grid of $64\times64$ candidates (Appendix~\ref{sec:implementation}). Exactly one is selected per image instance for training and prediction. We compare minimum-, middle-, and maximum-complexity selection with random selection. The complexity score aggregates directional filter responses, as in EIB-Net's selection procedure~\citep{eib2026}; the classifier receives the RGB patch, not those responses.

\begin{figure}[t]
\centering
\includegraphics[width=\linewidth]{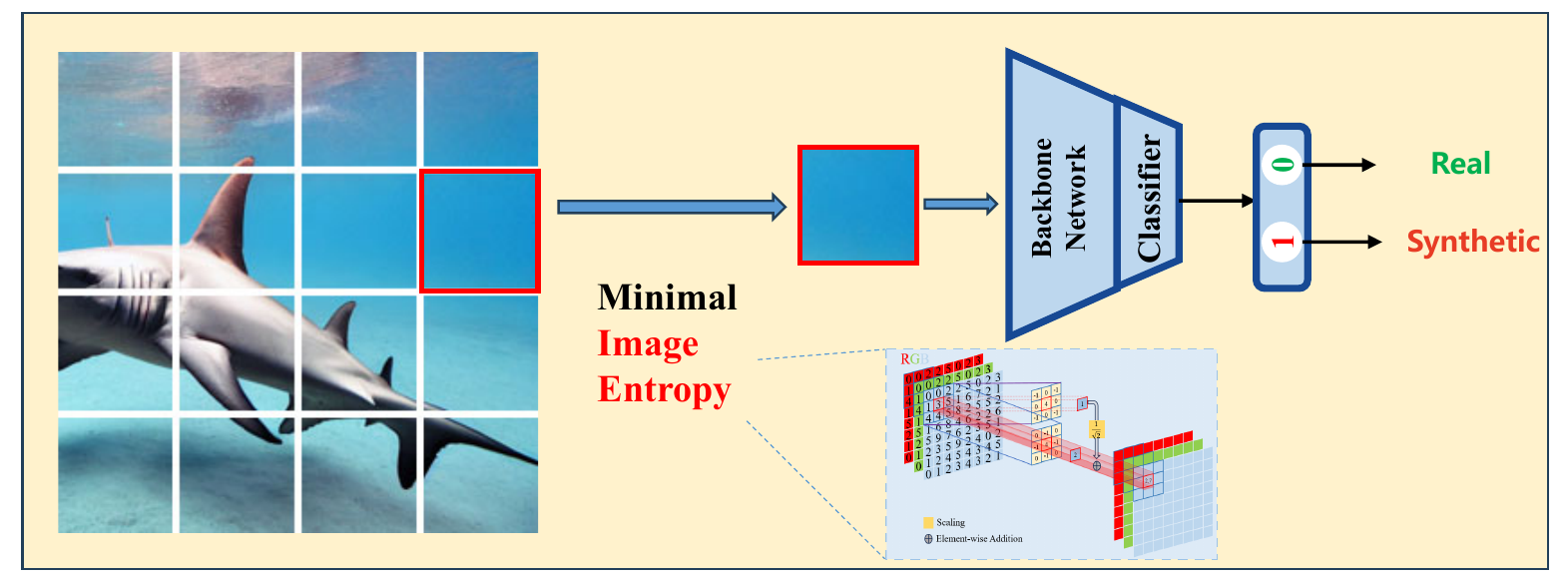}
\caption{The single-patch study, illustrated with minimum-complexity selection. We also test middle, maximum, and random selection. The original label \emph{Image Entropy} denotes a filter-based complexity score, not Shannon entropy. Each instance uses one RGB crop without aggregation.}
\label{fig:single-patch}
\end{figure}

Section~\ref{sec:results} shows that useful evidence is not confined to the minimum-complexity region. Random selection is competitive in the full-data study, while maximum-complexity selection is less accurate but still informative. This supports broadening spatial observations rather than assuming equal informativeness of all patches.

\subsection{Why multiple patches and separate losses?}
A single selected region can be ambiguous or weakly informative even when other regions are useful. Multiple crops reduce dependence on one choice. Coverage and supervision, however, are distinct: a multi-patch model can be trained to classify a pooled image representation or its constituent crops. We investigate the latter so that the objective explicitly evaluates every local prediction. Local decisions become the training units, while the image remains the evaluation unit.

\section{Patch-wise Supervision}
\label{sec:method}

\begin{figure*}[t]
\centering
\includegraphics[width=0.98\textwidth]{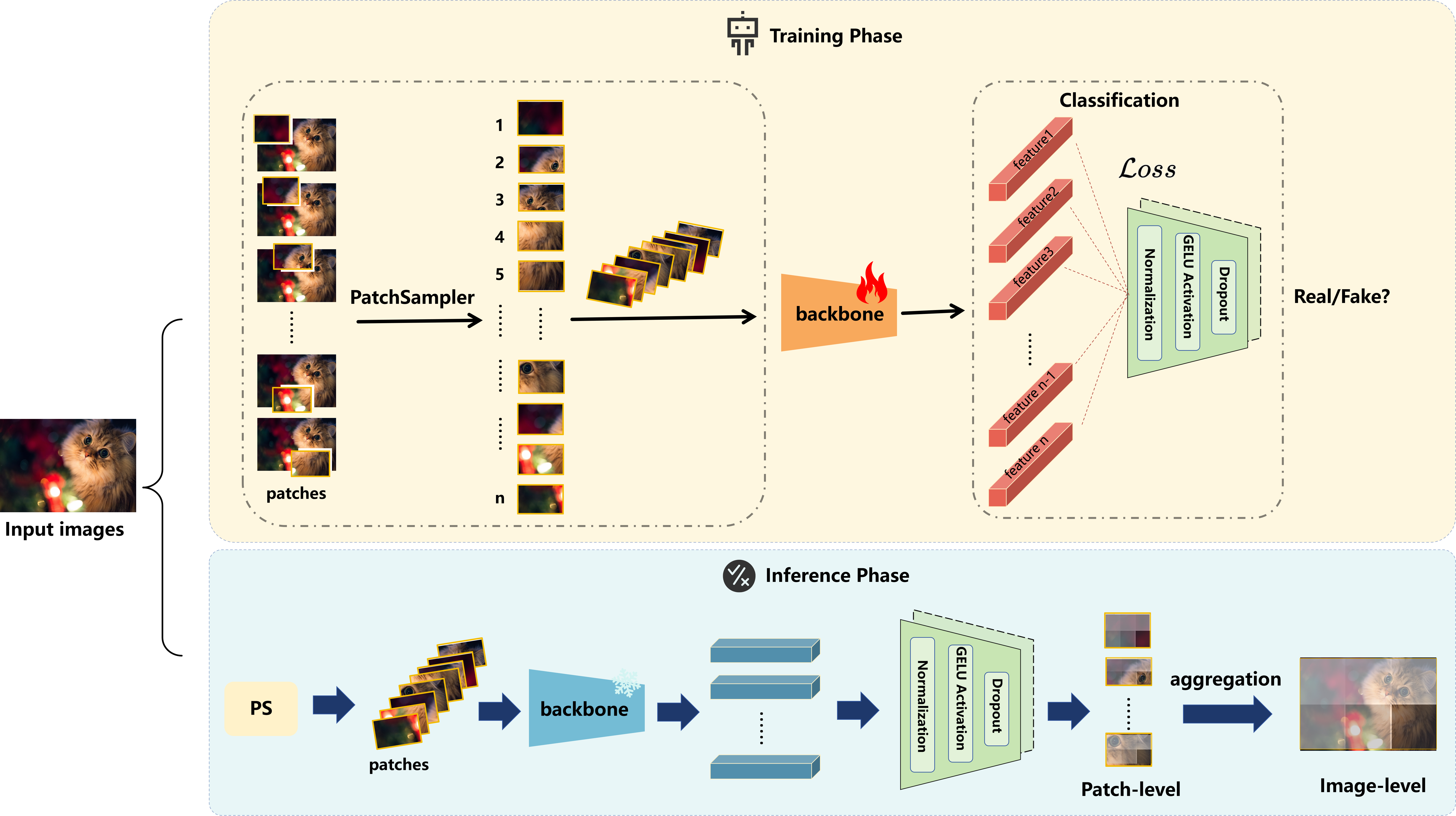}
\par\smallskip
\begin{minipage}{0.96\textwidth}\small\centering
\textbf{Training:} shared classifier per patch $\longrightarrow$
separate losses $\ell_1,\ldots,\ell_n$ $\longrightarrow$ mean loss.\\[2pt]
\textbf{Inference:} separate probabilities $q_1,\ldots,q_m$ $\longrightarrow$ mean probability $s(I)$.
\end{minipage}
\caption{Patch-wise supervision (PWS). All patches use the same backbone and head. The single head symbol denotes parameter sharing, not feature fusion: each patch produces its own logits and loss. Only probabilities are averaged during inference. The spatial score display visualizes predictions; it is not an input to the classifier.}
\label{fig:pipeline}
\end{figure*}

\subsection{Sampling and input representation}
Let $I_b$ have label $y_b\in\{0,1\}$, where $1$ denotes synthetic. The sampler extracts candidate regions and selects $n_b$ training patches. The candidate count, training count, and inference count $m_b$ are distinct. Selecting one of sixteen candidates still produces one training example; evaluating one crop of a multi-patch-trained detector is a different experiment.

We write the input path as
\begin{equation}
x_{bj}=T_{\rm patch}\bigl(C_j(T_{\rm image}(I_b))\bigr),
\label{eq:input}
\end{equation}
where $T_{\rm image}$ specifies image-level processing, $C_j$ extracts a crop, and $T_{\rm patch}$ specifies conversion, normalization, and any patch-level resizing. This separates the observed region from the tensor dimensions used by the backbone. Cropping before and after global downsampling are different operations.

The multi-patch recipe uses a nominal crop side of 64 pixels, stride 32, and a cap of 64 selected patches. Boundary-aligned windows cover the remaining right and bottom borders. When there are more candidates than the cap, a subset is selected without replacement. PWS requires no texture ranking or SRM residual transform. Historical input-size variants and the available implementation are specified in Section~\ref{sec:protocol} and Appendix~\ref{sec:implementation}.

\subsection{A shared classifier, one loss per patch}
A backbone $f_\theta$ and head $h_\theta$ produce two logits:
\begin{equation}
z_{bj}=h_\theta(f_\theta(x_{bj})),\qquad
q_{bj}=\operatorname{softmax}(z_{bj})_1.
\end{equation}
Every patch inherits its parent image's label. Patches from a mini-batch are concatenated, giving
\begin{equation}
\mathcal L_{\rm patch}=\frac{1}{K}\sum_{b=1}^{B}\sum_{j=1}^{n_b}
\ell(z_{bj},y_b),\qquad K=\sum_b n_b.
\label{eq:patchloss}
\end{equation}
Our implementation uses focal loss~\citep{focal},
\begin{equation}
\ell(z,y)=-\alpha(1-p_t)^\gamma\log p_t,\quad
p_t=\operatorname{softmax}(z)_y,
\end{equation}
with $\gamma=2$ and a common multiplier $\alpha=0.25$, rather than class-dependent weights. If patch counts vary, Equation~\ref{eq:patchloss} weights images in proportion to their patch counts. Inheriting image labels targets fully generated images; it is not a local manipulation annotation.

\subsection{Where supervision differs from pooling}
The image-level comparison first averages an image's patch logits:
\begin{equation}
\bar z_b=\frac{1}{n_b}\sum_jz_{bj},\qquad
\mathcal L_{\rm image}=\frac{1}{B}\sum_b\ell(\bar z_b,y_b).
\label{eq:jointloss}
\end{equation}
A nonlinear loss applied to pooled logits generally differs from the average of individual losses. Under Equation~\ref{eq:jointloss}, correct, confident predictions can compensate for poor predictions from other patches. Equation~\ref{eq:patchloss} instead penalizes those poor local predictions directly. This is the operational distinction in the supervision experiment.

The defining distinction is that each patch receives a classification loss before any within-image pooling. Parameters and optimization are shared; overlapping crops and training-time BatchNorm can introduce dependencies. No statistical independence assumption is required by this objective. Nor does averaging unordered logits itself reconstruct spatial layout. PWS changes the requirements on local predictions without assuming statistical independence or elimination of every semantic cue.

\subsection{Image-level prediction and score maps}
At inference, the same classifier processes $m_b$ patches:
\begin{equation}
s(I_b)=\frac{1}{m_b}\sum_{j=1}^{m_b}q_{bj},\qquad
\hat y_b=\mathbf{1}[s(I_b)\geq0.5].
\label{eq:inference}
\end{equation}
The average is over probabilities \emph{after} softmax, not logits. Evaluation uses fixed normalization statistics, permitting patch-wise chunking and score accumulation. Mapping predictions back to crop positions also reveals where the classifier is confident or uncertain. This visualizes decisions, not ground-truth artifact locations.

\subsection{Architecture and computation}
Explicit cropping constrains the input's visual field without truncating the feature extractor. We evaluate ResNet-18/50~\citep{resnet}, Xception~\citep{xception}, and Swin-T~\citep{swin}. Each backbone still requires a valid input-size configuration, explicitly set in the available Swin wrapper. The method is simple to implement, but its cost depends on input size and patch count. Multiple crops are not automatically faster than one whole-image pass, and more crops need not improve accuracy monotonically.

%% file: sections/experiments.tex
\section{Experiments}
\label{sec:experiments}
\label{sec:protocol}

Our experiments address three questions: how much detection evidence survives in a single local crop; whether patch-wise learning transfers across generators and backbone architectures; and how supervision, spatial resolution, and inference coverage affect the result. We also examine image processing and difficult evaluation images to identify the approach's limitations.

\subsection{Evaluation settings}
\label{sec:setup}

\paragraph{Datasets.}
The principal multi-patch experiments use GenImage~\citep{genimage} and AIGCDetectBenchmark~\citep{patchcraft}, abbreviated AIGCD. For GenImage, the training generator is SD~v1.4 and evaluation covers eight generator subsets, including the training generator and BigGAN. For AIGCD, our models are trained on ProGAN and evaluated on the sixteen subsets listed in Appendix~\ref{app:full-results}. These benchmark means include the source generator; they are not exclusively unseen-generator averages. The two collections provide complementary diffusion-source and GAN-source settings. Single-patch studies use the DIFF facial-image setting~\citep{diff} and a separate LSUN-to-ImageNet transfer setting from DiffusionForensics~\citep{dire}. Chameleon~\citep{aide} tests performance on a more challenging collection of real and generated images.

\paragraph{Models and metrics.}
We use ResNet-18, ResNet-50~\citep{resnet}, Xception~\citep{xception}, and Swin-T~\citep{swin} to examine the same learning recipe with convolutional and Transformer backbones. Unless stated otherwise, the multi-patch setting uses $64\times64$ crops and up to 64 inference patches. Accuracy is computed from the mean patch probability with threshold 0.5; average precision (AP) measures image-level ranking. Reported benchmark means average generator subsets rather than random seeds. For Chameleon, we show both class accuracies and balanced accuracy to expose class-dependent errors.

\paragraph{Input processing.}
The earlier manuscript describes native-resolution cropping for the default multi-patch setting, followed by backbone-specific transforms, and a separate resolution-aligned control. The retained source instead applies a $299\times299$ source resize annotated as an ablation, then passes normalized $64\times64$ patches directly to the feature extractors. Neither this snapshot nor the public implementation certifies the complete preprocessing history of every reported run. We distinguish source resizing from post-crop resizing and retain the historical settings separately; Appendix~\ref{sec:input-variants} gives the version details.

\paragraph{Comparison and model selection.}
\label{sec:selection}
This paper reports the existing experiments without new training. External-method scores are retained from the original comparison tables, with attribution, rather than presented as newly rerun baselines. Historical checkpoint and run selection was not uniform: validation-based selection was used where available, but some results were chosen using target evaluation performance. In particular, the ResNet-50 GenImage result of 95.40 follows a scan maximizing the eight-subset evaluation average; the SD~v1.4-trained Xception Chameleon result of 71.29 is the best checkpoint in a Chameleon evaluation scan. Other rows' complete selection histories have not been recovered. Because selection histories are not consistently matched or fully recovered, the reported patch-versus-whole-image differences do not establish the direction or magnitude of gains under a common evaluation protocol. Superscript T marks documented target-selected results; unmarked results from our experiments have incompletely recovered selection histories, rather than verified source-only selection. External-method rows retain their original attribution. Target-based selection can make reported scores optimistic even without using those images for gradient updates. We retain the comparisons to show the observed behavior, but do not interpret their differences as an unbiased state-of-the-art ranking or as statistically established effect sizes. Known numerical inconsistencies are identified locally and in Appendix~\ref{sec:numerical-notes}.

For the same Xception run, a separate evaluation of its saved best checkpoint reports 68.03 on Chameleon, compared with 71.29 from the Chameleon checkpoint sweep. This 3.26-point difference illustrates sensitivity to checkpoint selection, not an unbiased estimate of test-selection bias (Appendix~\ref{sec:selection-example}).

\subsection{From one local observation to multiple patches}
\label{sec:spd_exp}

\paragraph{Useful evidence is not limited to the simplest crop.}
The documented single-patch study divides each $256\times256$ image into sixteen non-overlapping $64\times64$ candidates, using exactly one per training and evaluation instance. Table~\ref{tab:single} compares minimum-, middle-, and maximum-complexity selection with random selection. This is a single-patch detection baseline (SPD), not the complete EIB-Net model with its information-bottleneck module.

\begin{table}[t]
\centering\small
\caption{Single-patch selection on the full-data DIFF setting. One crop occupies 6.25\% of the resized image canvas. This is an input-area fraction, not a measured compute ratio.}
\label{tab:single}
\begin{tabular}{lc}
\toprule
Input / selection & Accuracy (\%)\\
\midrule
Whole image & 99.01\\
Single patch, minimum complexity & 98.66\\
Single patch, middle complexity & 97.93\\
Single patch, maximum complexity & 94.91\\
Single patch, random & 98.78\\
\bottomrule
\end{tabular}
\end{table}

Random selection reaches 98.78\%, close to the whole-image result of 99.01\% and slightly above minimum-complexity selection. Even maximum-complexity crops retain substantial detection ability, although their 94.91\% accuracy is clearly lower. The result motivates using local RGB content without requiring one privileged texture range. It does not imply that every region is equally informative. In particular, committing to one selected crop can discard useful evidence elsewhere in the image. Multi-patch inference addresses this coverage issue while maintaining patch-level training predictions.

\begin{figure*}[t]
\centering
\includegraphics[width=0.83\textwidth]{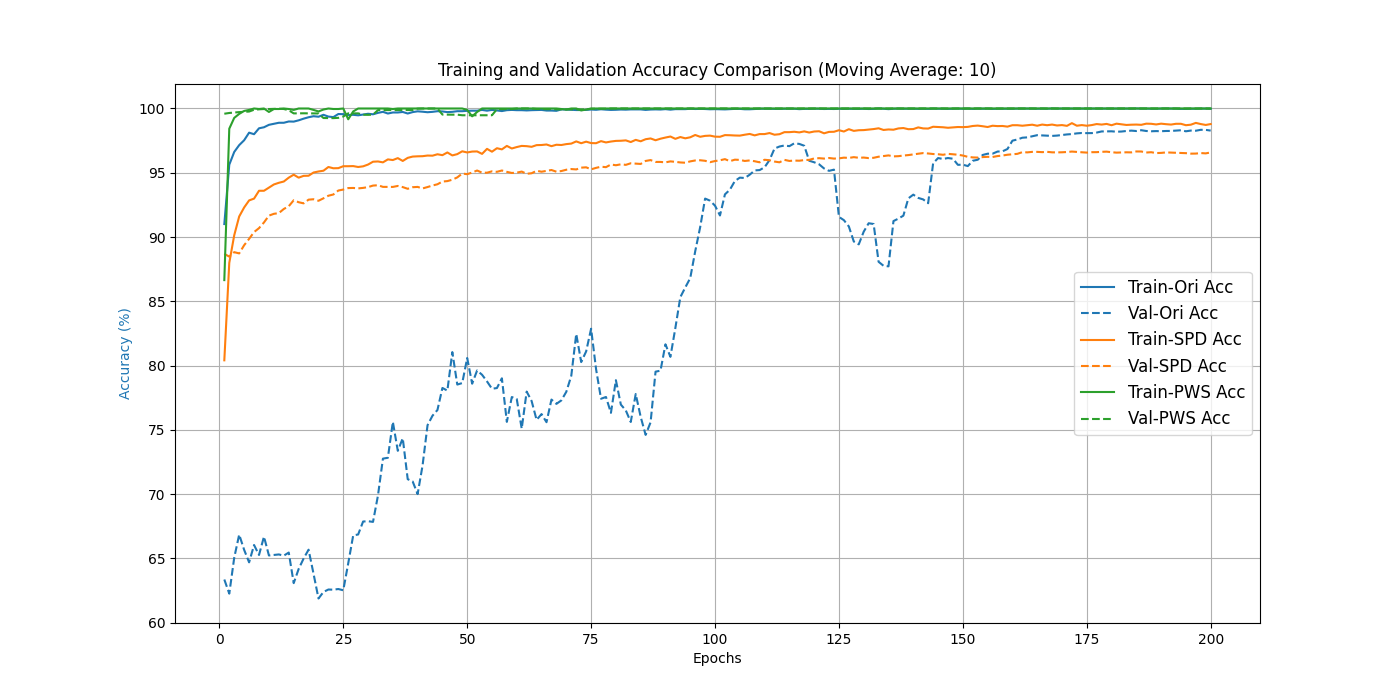}
\caption{Recorded training and validation accuracy over 200 epochs, shown with the original 10-epoch moving-average visualization. Ori denotes whole-image learning, SPD the single-patch study, and PWS patch-wise supervision. These trajectories illustrate observed training behavior, not a causal test of the learned features.}
\label{fig:training}
\end{figure*}

\paragraph{Training behavior.}
Figure~\ref{fig:training} compares the recorded training and validation trajectories for whole-image, single-patch, and multi-patch models. The whole-image run has a larger train--validation gap, while the patch-based runs track validation performance more closely. This observation complements the final scores: local-input learning can change optimization behavior as well as the evaluation result. The curves alone do not identify which visual cues the models use or prove the removal of semantic bias. Further single-patch selection, training-data, and transfer results are retained in Appendix~\ref{app:single-transfer}.

\subsection{Performance across generators and backbones}
\label{sec:results}

\input{tables/main_genimage}
\input{tables/main_aigcd}

\paragraph{GenImage.}
Table~\ref{tab:GenImage} presents the complete eight-generator comparison. The reported whole-image and PWS mean accuracies are, respectively, 72.63 and 93.97 for ResNet-18, 75.06 and 95.40 for ResNet-50, 79.42 and 95.21 for Xception, and the reported 77.84 and 93.58 for Swin-T. Whole-image models are already highly accurate on SD~v1.4 and SD~v1.5, so much of the reported difference occurs on more distant generator subsets. For example, the ResNet-50 whole-image and PWS values are 61.25 and 95.61 on Midjourney, and 59.67 and 94.82 on VQDM.

In the reported table, PWS has a higher mean than the corresponding whole-image baseline for each of the four backbones, although this ordering does not hold for every generator subset: Swin-T has 78.25 for whole-image input and 78.08 for PWS on GLIDE. These comparisons do not establish the direction or magnitude of gains under a common selection protocol (Section~\ref{sec:selection}). External-method rows provide context, rather than an unbiased method ranking.

\paragraph{AIGCD.}
Table~\ref{tab:AIGCD} summarizes the second source setting, with full per-generator accuracy and AP in Appendix~\ref{app:full-results}. PWS reaches mean accuracies of 88.41--93.43 across the four backbones. Swin-T and Xception obtain 93.43 and 93.02, respectively, compared with the listed 92.77 for AIDE and 89.31 for PatchCraft. The smaller ResNet models do not exceed all external baselines, which is important when interpreting the recipe as a portable design rather than a universal performance guarantee.

The detailed results also show a persistent weakness on GauGAN: PWS accuracy ranges from 57.08 for ResNet-18 to 76.80 for Swin-T. At the same time, average AP is appreciably higher than fixed-threshold accuracy; for example, ResNet-50 obtains 95.11 AP versus 89.35 accuracy. Thus, a useful score ranking need not yield a well-calibrated threshold across generators. These observations motivate examining both the local feature representation and how patch scores are calibrated and aggregated.

\subsection{What changes when the supervision unit changes?}
\label{sec:ablation_supervision}

Both objectives in Equations~\ref{eq:patchloss} and~\ref{eq:jointloss} process cropped inputs. Their difference is whether the loss is applied to every prediction or to the mean logits for an image. Table~\ref{tab:supervision} shows higher GenImage means for patch-level supervision on all three evaluated backbones: 95.40 versus 72.40 for ResNet-50, 95.21 versus 80.19 for Xception, and 93.58 versus 77.70 for Swin-T.

\begin{table}[t]
\centering\small
\caption{Supervision granularity on GenImage, mean accuracy (\%). Image-level supervision averages patch logits before the loss; PWS applies a loss to each patch. Per-generator values appear in Appendix~\ref{app:ablation-details}. T marks the documented target-selected result; other entries have incompletely recovered selection histories.}
\label{tab:supervision}
\begin{tabular}{lrr}
\toprule
Backbone & Image-level loss & Patch-level loss\\
\midrule
ResNet-50 & 72.40 & 95.40$^{\mathrm{T}}$\\
Xception & 80.19 & 95.21\\
Swin-T & 77.70 & 93.58\\
\bottomrule
\end{tabular}
\end{table}

The distinction is practically consequential: requiring each crop to predict the label discourages a model from satisfying the objective only through a few easy regions. Nevertheless, the numerical comparison does not isolate semantic bias as the causal explanation. The objectives also differ in optimization and example weighting, while shared parameters and batch statistics remain. The supported conclusion is that explicit local input and per-patch loss should be considered together; simply introducing patches does not specify the training problem.

\subsection{Resolution, patch size, and inference coverage}
\label{sec:ablation}

\paragraph{Cropping versus access to native-resolution detail.}
A default crop can preserve local information that global resizing removes. The aligned experiment first resizes each image to the whole-image baseline's canvas, then extracts patches. Table~\ref{tab:aligned} compares this setting with whole-image and default-patch processing.

\begin{table}[t]
\centering\small
\caption{Resolution-aligned GenImage comparison, mean accuracy (\%). Aligned patches are extracted from a 224-pixel canvas for ResNet-50 and a 299-pixel canvas for Xception. T marks the documented target-selected result; other entries have incompletely recovered selection histories.}
\label{tab:aligned}
\setlength{\tabcolsep}{4pt}
\begin{tabular}{lrrr}
\toprule
Backbone & Whole image & Aligned PWS & Default PWS\\
\midrule
ResNet-50 &75.06&85.21&95.40$^{\mathrm{T}}$\\
Xception &79.42&85.43&95.21\\
\bottomrule
\end{tabular}
\end{table}

The reported aligned-PWS means exceed the whole-image means by 10.15 percentage points for ResNet-50 and 6.01 points for Xception; the reported default-PWS means are higher still. These are descriptive differences between historical results. Unresolved preprocessing and selection histories prevent interpreting them as isolated effects of source resolution or supervision granularity.

\paragraph{Patch size.}
The $64\times64$ setting has the highest mean in the Xception comparison on both benchmarks (Table~\ref{tab:size}). Smaller crops reduce the available spatial context, while larger crops alter the balance between local detail and broader image structure. The observed optimum is therefore a useful default for these experiments, not a universal property of generated images. Per-generator changes in Appendix~\ref{app:ablation-details} also show why a single mean should not substitute for examining individual distributions.

\begin{table}[t]
\centering\small
\caption{Xception patch-size comparison, accuracy (\%). The 128-pixel entries follow the detailed per-generator tables; the earlier main-text table interchanged their dataset columns.}
\label{tab:size}
\begin{tabular}{lrr}
\toprule
Patch size & GenImage & AIGCD\\
\midrule
$32\times32$ &94.30&91.77\\
$64\times64$ &95.21&93.02\\
$128\times128$ &91.45&91.54\\
\bottomrule
\end{tabular}
\end{table}

\paragraph{Inference coverage.}
In the reported inference patch-count study (Table~\ref{tab:count}), the mean accuracies are 87.33, 92.45, and 93.02 for one, sixteen, and sixty-four crops, respectively. The reported means are not monotonic as the count increases further. This study concerns inference coverage for a patch-trained classifier, rather than the separate single-patch training experiment in Table~\ref{tab:single}.

\begin{table}[t]
\centering\small
\caption{Inference patch-count study on AIGCD. WFR and mean accuracies are percentages. The original elapsed-time observations are retained, but their workload and timing boundary are unspecified; they should not be converted to per-image latency or compared with other implementations.}
\label{tab:count}
\begin{tabular}{rrrr}
\toprule
Test patches & WFR & Mean & Recorded time (s)\\
\midrule
1&92.95&87.33&6.04\\
4&92.95&91.75&6.19\\
16&96.05&92.45&6.69\\
64&96.45&93.02&9.06\\
128&96.40&92.67&16.73\\
256&96.65&92.70&30.08\\
\bottomrule
\end{tabular}
\end{table}

These results suggest a coverage--cost trade-off rather than a rule that more crops are always better. Selecting fewer informative patches, weighting unreliable predictions, and accounting for overlapping evidence are natural extensions, but are not required by the simple mean-aggregation method evaluated here.

\subsection{Processing sensitivity and difficult images}
\label{sec:robustness}

\input{tables/main_perturbations}
\input{tables/main_chameleon}

\paragraph{Image processing.}
Table~\ref{tab:perturbations} compares detectors under JPEG compression, half-resolution downsampling, and Gaussian blur. The PWS Swin-T result decreases from the clean 93.43 to 86.83 under downsampling and 87.18 under blur, drops of 6.60 and 6.25 percentage points. JPEG produces a larger decrease to 80.71. The table does not establish resistance to severe compression: its JPEG quality factor is not identified unambiguously in the retained records. The individual-generator accuracy and AP tables are included in Appendix~\ref{app:full-results}.

The relative ordering in these recorded results is encouraging, but the clean-to-processed decline exposes an important limitation of local detection. Resizing and compression modify precisely the fine-scale statistics on which a local classifier may depend. This can affect real photographs as well as synthetic images. Robust local evidence should therefore be studied alongside realistic acquisition and post-processing pipelines, rather than equated with invariance to such operations.

\paragraph{Chameleon.}
The class breakdown in Table~\ref{tab:chameleon} shows differences in reported performance and substantial remaining failure. The listed AIDE models recognize real images well but detect relatively few synthetic examples. PWS has higher reported synthetic-class accuracy in these comparisons, reaching 37.58 with SD~v1.4-trained Xception while retaining 96.63 real-class accuracy. However, this still means that most synthetic images are missed at the fixed threshold. Its 71.29 overall accuracy corresponds to only 67.11 balanced accuracy and, as noted in Section~\ref{sec:selection}, uses a target-selected checkpoint. The result describes a target-selected operating point in this difficult setting; it does not establish a gain under an independent selection protocol or reliable open-world detection.

Taken together, the experiments favor treating local RGB classification and its supervision unit as meaningful design choices. They also show why the scope of the conclusion matters: useful local signals need not exist in every crop, transfer uniformly to every generator, survive all processing, or support a single universal decision threshold.

%% file: tables/main_genimage.tex
\begin{table*}[t]
\centering\footnotesize
\caption{GenImage accuracy (\%) across eight generator subsets, with SD~v1.4 as the training source. Whole-image and PWS rows use the indicated backbone; external-method values are retained from the original comparison. Means include SD~v1.4. T marks documented target-based selection; other whole-image/PWS rows have incompletely recovered selection histories. Selection-policy differences are described in Section~\ref{sec:selection}. $\dagger$ marks an inconsistency between the reported mean and the displayed subset scores (Appendix~\ref{sec:numerical-notes}).}
\label{tab:GenImage}
\label{tab:backbones}
\setlength{\tabcolsep}{4pt}
\begin{tabular}{llrrrrrrrrr}
\toprule
Backbone / detector & Input / method & Midj. & SD1.4 & SD1.5 & ADM & GLIDE & Wukong & VQDM & BigGAN & Mean\\
\midrule
\multicolumn{2}{l}{DIRE~\citep{dire}} &60.20&99.90&99.80&50.90&55.00&99.20&50.10&50.20&70.66\\
\multicolumn{2}{l}{GenDet~\citep{gendet}} &89.60&96.10&96.10&58.00&78.40&92.80&66.50&75.00&81.56\\
\multicolumn{2}{l}{PatchCraft~\citep{patchcraft}} &79.00&89.50&89.30&77.30&78.40&89.30&83.70&72.40&82.30\\
\multicolumn{2}{l}{AIDE~\citep{aide}} &79.38&99.74&99.76&78.54&97.82&98.65&80.26&66.89&86.88\\
\midrule
ResNet-18&Whole image&62.33&99.75&99.31&56.00&60.75&96.17&55.67&51.08&72.63\\
&PWS&89.06&99.89&99.89&89.58&99.62&99.88&85.00&88.85&93.97\\
\addlinespace[2pt]
ResNet-50&Whole image&61.25&99.92&99.81&57.83&69.58&98.67&59.67&53.75&75.06\\
&PWS$^{\mathrm{T}}$&95.61&99.91&99.91&88.74&98.47&99.87&94.82&85.88&95.40\\
\addlinespace[2pt]
Xception&Whole image&60.58&99.92&99.62&65.75&84.92&98.58&67.67&58.33&79.42\\
&PWS&84.34&99.98&99.92&83.88&95.73&99.98&99.42&98.38&95.21\\
\addlinespace[2pt]
Swin-T&Whole image&62.67&99.92&99.56&61.92&78.25&98.00&62.67&58.08&77.84$^\dagger$\\
&PWS&87.62&99.95&99.88&94.93&78.08&99.93&98.52&89.69&93.58\\
\bottomrule
\end{tabular}
\end{table*}

%% file: tables/main_aigcd.tex
\begin{table}[t]
\centering\small
\caption{AIGCD mean accuracy and AP (\%) across sixteen subsets. PWS models are trained on ProGAN; their complete selection histories have not been recovered (Section~\ref{sec:selection}). External rows preserve the original comparison, including DIRE's separate G/D variants. A dash denotes an unavailable metric, not zero. Per-generator values are in Appendix~\ref{app:full-results}; $\dagger$ is explained in Appendix~\ref{sec:numerical-notes}.}
\label{tab:AIGCD}
\begin{tabular}{lrr}
\toprule
Detector & Accuracy & AP\\
\midrule
CNNSpot~\citep{wang2020}&70.78&80.41\\
LNP~\citep{lnp}&83.84&91.58\\
LGrad~\citep{lgrad}&75.34&81.91\\
DIRE-G~\citep{dire}&68.68&78.78\\
DIRE-D~\citep{dire}&71.53&--\\
UnivFD~\citep{ojha2023}&--&91.74$^\dagger$\\
PatchCraft~\citep{patchcraft}&89.31&95.83\\
NPR~\citep{npr}&82.91&--\\
AIDE~\citep{aide}&92.77&--\\
\midrule
PWS (ResNet-18)&88.41&93.97\\
PWS (ResNet-50)&89.35&95.11\\
PWS (Xception)&93.02&97.69$^\dagger$\\
PWS (Swin-T)&93.43&98.48\\
\bottomrule
\end{tabular}
\end{table}

%% file: tables/main_perturbations.tex
\begin{table}[t]
\centering\small
\caption{AIGCD mean accuracy (\%) after image processing. Downsampling uses ratio 0.5 and blur uses $\sigma=1$ in the experiment description. JPEG results are retained from the original report. The corresponding perturbation-evaluation script has not been located in the retained code, so the quality factor cannot be verified and this column cannot currently be reproduced from the available materials. AIDE's downsampling value was not reported.}
\label{tab:perturbations}
\begin{tabular}{lrrr}
\toprule
Detector & JPEG & Downsample & Blur\\
\midrule
CNNSpot~\citep{wang2020}&64.03&58.85&68.39\\
LNP~\citep{lnp}&53.74&63.55&67.20\\
LGrad~\citep{lgrad}&51.54&60.86&71.29\\
DIRE-G~\citep{dire}&66.57&56.09&68.92\\
DIRE-D~\citep{dire}&70.27&62.26&70.69\\
UnivFD~\citep{ojha2023}&74.25&70.87&72.98\\
PatchCraft~\citep{patchcraft}&72.20&78.38&75.88\\
AIDE~\citep{aide}&75.54&--&81.88\\
\midrule
PWS (Swin-T)&80.71&86.83&87.18\\
\bottomrule
\end{tabular}
\end{table}

%% file: tables/main_chameleon.tex
\begin{table*}[t]
\centering\small
\caption{Chameleon accuracy (\%). Syn. and Real report class-wise accuracy; BAcc is their arithmetic mean, calculated here from the displayed values. AIDE rows are retained from the original comparison. T marks documented target-based selection; other PWS rows have incompletely recovered selection histories. The SD~v1.4/Xception PWS row uses a checkpoint selected on Chameleon (Section~\ref{sec:selection}).}
\label{tab:chameleon}
\begin{tabular}{lllrrrr}
\toprule
Training source & Detector & Backbone & Syn. & Real & Overall & BAcc\\
\midrule
ProGAN&AIDE~\citep{aide}&ResNet-50&0.63&98.46&56.45&49.55\\
ProGAN&PWS&ResNet-50&28.79&98.69&68.70&63.74\\
ProGAN&PWS&Xception&28.15&97.92&67.98&63.04\\
\midrule
SD~v1.4&AIDE~\citep{aide}&ResNet-50&16.82&94.38&61.10&55.60\\
SD~v1.4&PWS&ResNet-50&27.00&93.59&65.02&60.30\\
SD~v1.4&PWS$^{\mathrm{T}}$&Xception&37.58&96.63&71.29&67.11\\
\bottomrule
\end{tabular}
\end{table*}

%% file: sections/discussion.tex
\section{Discussion and Limitations}
\label{sec:discussion}

\paragraph{What the results support.}
The study supports treating spatial input and supervision granularity as useful design choices. A standard RGB classifier can learn from small regions; multiple decisions provide broader coverage; and patch-level training performs favorably across the tested backbones in the reported comparisons. This makes a simple alternative to specialized forensic representations worth studying, without assuming every architecture or distribution will benefit.

\paragraph{Local evidence is not a uniquely identified mechanism.}
Restricting the visual field reduces access to full-image composition, but crops can still contain semantic cues, compression differences, and other provenance shortcuts. The supervision and resolution experiments help distinguish design choices; they do not uniquely attribute improvements to neighboring-pixel generation patterns. Likewise, gradient co-occurrence describes input statistics, not the network's learned features. Broader crop coverage, the augmentation effect of training on multiple local views, and inference-time averaging of correlated predictions may also contribute to the reported behavior; the present comparisons do not isolate their contributions from those of spatial context and supervision granularity. Stronger causal claims require controls beyond this study.

\paragraph{Coverage and patch labels.}
Uniform averaging gives clear, ambiguous, and corrupted regions equal weight. Overlap can introduce redundancy. Adaptive selection and confidence-weighted pooling are possible extensions, but their benefit is not established here. Furthermore, inheriting an image label for every crop targets fully synthesized images. Partial edits and mixed real/synthetic compositions require different treatment of local labels and are outside the demonstrated scope.

\paragraph{Evaluation boundaries.}
The selection practices in Section~\ref{sec:selection} limit score interpretation: evaluation-selected results can be optimistic, and cross-method selection budgets are not matched. The comparisons characterize reported behavior rather than establish an unbiased state-of-the-art ranking. The experiments also do not provide systematic multi-seed estimates. Difficult-image failures and processing sensitivity show that high benchmark averages alone do not establish reliability in deployment.

\paragraph{Practical use and release.}
A detector score is supporting evidence, not proof of image origin. False positives can cause harmful accusations; high real-image accuracy can also coexist with low synthetic-image recall. The public reference implementation is available at \url{https://github.com/LF-Jade/look-closer}. It covers multi-patch PWS sampling, patch-level training, and image-level evaluation, with configuration and historical-version limitations documented in the repository. The historical single-patch selection implementation is not included, and no detector checkpoints trained for this study are provided. The release is not a guarantee that every reported score can be reproduced.

\section{Conclusion}
We studied a simple route from single-patch evidence to multi-patch detection: explicitly crop RGB regions, train a shared classifier with a loss for each patch, and average probabilities at inference. The experiments motivate looking beyond one specially selected region and report higher GenImage means for PWS than for the whole-image baselines across four tested backbones, with selection histories that limit conclusions under a common evaluation protocol. Supervision, resolution, and coverage studies characterize the design space, while processing sensitivity and difficult-image failures define its limits. The work offers a practical recipe and observations for further investigation of local supervision, without requiring a new feature extractor or claiming that local evidence alone solves image authenticity.

%% file: sections/implementation.tex
\section{Implementation Details}
\label{sec:implementation}

\subsection{Patch extraction, model, and optimization}
The multi-patch implementation reads RGB images and binary image labels, extracts patches, and repeats the image label for each patch. A collated batch concatenates these patches and retains their source-image counts. The shared model returns two logits per patch. Training averages the patch-level focal losses; evaluation groups synthetic-class softmax probabilities by source image, averages them, and uses a decision threshold of 0.5.

\begin{center}
\small
\captionof{table}{Multi-patch implementation details. Input resizing is version-dependent, as discussed below; the remaining entries describe the saved training and evaluation path.}
\label{tab:implementation}
\begin{tabularx}{\textwidth}{@{}p{0.23\textwidth}X@{}}
\toprule
Component & Implementation\\
\midrule
Sampling & Nominal patch side 64, stride 32, capped candidate subset, with right/bottom boundary-aligned windows.\\
Normalization & ImageNet mean $(0.485,0.456,0.406)$ and standard deviation $(0.229,0.224,0.225)$.\\
Backbone inputs & ResNet and Xception receive the supplied patch tensor without internal interpolation; Swin explicitly sets its input size to 64.\\
Classification head & Linear projection to 512, BatchNorm1d, GELU, dropout 0.5, and a two-logit linear layer.\\
Loss & Mean patch-level focal loss with common multiplier $\alpha=0.25$ and focusing exponent $\gamma=2$.\\
Optimizer & AdamW, configured learning rate, weight decay 0.01, and mixed-precision training.\\
Image-level control & Average logits within each image before applying the classification loss.\\
Inference & Average synthetic-class softmax probabilities within each source image.\\
\bottomrule
\end{tabularx}
\end{center}

The sampler uses a seed and provides an epoch-offset parameter, but the saved training loop does not update that offset. Accordingly, epoch-wise resampling and translation jitter are not assumed here. Boundary-aligned windows also differ from cyclic padding. These details do not change the distinction between averaging individual patch losses and applying a loss after image-level aggregation.

\paragraph{Representative configurations.}
The ResNet-50 GenImage configuration specifies a patch side of 64, stride 32, a cap of 64 patches, an image batch size of 32, learning rate 0.005, seed 42, 30 training epochs, and checkpoint saving every 2000 iterations. The Xception/SD~v1.4 configuration associated with the Chameleon evaluation uses the same sampling and batch settings but specifies 20 epochs. The number of patches actually supplied by each image determines the effective patch batch size. These are run-specific settings, not a claim that every experiment used a common schedule. The experiments were conducted on RTX 4090 servers.

\subsection{Input resolution and historical variants}
\label{sec:input-variants}
Two spatial operations must be distinguished: resizing the \emph{source image before cropping}, and resizing an extracted \emph{patch before the backbone}. In the saved multi-patch source, a $64\times64$ patch is passed directly to ResNet, Xception, or Swin after normalization. The model wrappers do not silently enlarge it to $224\times224$ or $299\times299$. In particular, a pretrained model's identifier is not evidence that its default input transform is applied.

The earlier manuscript describes native-resolution cropping for the default multi-patch setting, followed by the same backbone-specific transforms as the paired whole-image baseline. Its resolution-aligned control instead resizes the source to a fixed $224\times224$ or $299\times299$ canvas before cropping and then enlarges the extracted patches to the backbone input size. The retained sampler unconditionally applies a $299\times299$ resize before cropping, explicitly annotated as an ablation, whereas the retained model path consumes $64\times64$ patches directly. The snapshot therefore documents a resize-before-crop ablation path but does not establish an exact implementation match to either complete manuscript pipeline or identify the transform version used for every historical result. We retain the reported settings as distinct historical comparisons, without treating their differences as isolated effects of resolution. The public implementation defaults to native-resolution cropping with direct-$64$ input; this is an explicit release configuration, not a claim to have recovered the full historical pipeline.

\subsection{Checkpoint counters and selection example}
\label{sec:selection-example}
When the save interval is positive, the training loop increments a checkpoint counter every specified number of batch steps. A filename such as \texttt{epoch146.pth} need not mean the 146th completed training epoch. For the representative ResNet-50 schedule, 30 epochs with image batch size 32 and checkpoint saving every 2000 steps are consistent with the 151 recorded checkpoints.

For the Xception example in Section~\ref{sec:selection}, the GenImage log selects checkpoint 43 by mean accuracy across eight generators. A subsequent evaluation of the run's \texttt{best\_epoch.pth} reports 68.03 on Chameleon; the Chameleon sweep also reports 68.03 for checkpoint 43 and selects checkpoint 16 at 71.29. This is a log-level association, not an immutable weight-identity record or proof of prospective checkpoint freezing before inspecting Chameleon. GenImage cross-generator selection is not source-generator-only validation. The recorded difference demonstrates selection sensitivity, but does not isolate test-selection optimism from genuine differences between checkpoints.

\subsection{Single-patch experiments and relation to EIB-Net}
The single-patch baseline (SPD) uses exactly one selected region per image instance. Its documented default first resizes the image to a $256\times256$ canvas, forms a $4\times4$ grid, and chooses one $64\times64$ region. The original single-patch description includes backbone-dependent transforms after selection; this path is distinct from the multi-patch implementation above. The documented SPD training setup uses ImageNet initialization, seed 3407, and JPEG augmentation with probability 0.5 and quality 95. CNN runs use SGD with momentum 0.9 and initial learning rate $10^{-2}$; Transformer runs use AdamW with initial learning rate $10^{-4}$ and weight decay $10^{-4}$.

Minimum-, middle-, and maximum-complexity selection use the local variation statistic developed in our earlier research~\citep{eib2026}. It sums absolute directional Laplacian-style responses and is not Shannon entropy. Selection changes which RGB crop is supplied; the statistic is not itself the detector input. Random selection uses no complexity score. SPD is a single-patch baseline, not the complete EIB-Net architecture: EIB-Net additionally includes a variational information bottleneck. The two studies share research origins, so their scores should not be treated as independent replications or as a controlled bottleneck ablation.

\subsection{Numerical conventions and corrections}
\label{sec:numerical-notes}
This paper reorganizes existing experiments without adding training runs. Accuracy and average precision retain the original percentage scale. Balanced accuracy, where shown, is the arithmetic mean of the two class accuracies. The detailed 128-pixel patch-size rows average to 91.4525 on GenImage and 91.53625 on AIGCDetectBenchmark; we correct the interchange of these means in the earlier summary table.

Several reported summaries do not match their displayed component rows: GenImage Swin-T whole-image accuracy averages to 77.63375 (77.63), rather than 77.84; AIGCD Xception AP averages to 97.67875 (97.68), rather than 97.69; UnivFD AIGCD AP averages to 91.115625 (91.12), rather than 91.74; and PWS Swin-T blur AP averages to 91.6775 (91.68), rather than 91.35. The historical summaries are retained and marked because the tables alone cannot identify whether a component or its mean was transcribed incorrectly. The $64\times64$ GenImage Xception patch row averages to 95.20375 rather than 95.21; this smaller difference can arise from averaging before rounding. The DIRE-D/ProGAN accuracy entry is corrected from the malformed ``52..75'' to 52.75, as recorded in two other saved copies of the same table. Its 16 entries sum to 1144.55 and average to 71.534375, consistent with the reported 71.53. These issues do not reverse the reported whole-image versus patch-wise comparison, but should be resolved from raw results before using small numerical differences to rank models.

Low-data rows with conflicting subset labels and timing values without a complete measurement protocol are annotated where reported. They are not used to claim a precisely quantified sample-complexity or runtime advantage. Checkpoint-selection qualifications are stated with the experimental protocol in Section~\ref{sec:protocol}.

%% file: sections/appendix_tables.tex
\section{Full Per-Generator Results}
\label{app:archival-tables}
\label{app:full-results}

The following tables expand the benchmark comparisons and perturbation studies.
PWS denotes the multi-patch method, whereas
SPD denotes a single-patch input and is not the complete EIB-Net model.
All values are percentages. The AIGCDetectBenchmark tables list 16 generator
subsets; their reported means include ProGAN, the training source for PWS.
They are not averages over 16 unseen generators.

The external detector scores are retained from the comparison tables used in
the original study, not newly reproduced for this version.
Relevant methods include CNNSpot~\citep{wang2020},
DIRE~\citep{dire}, UnivFD~\citep{ojha2023}, PatchCraft~\citep{patchcraft},
NPR~\citep{npr}, and AIDE~\citep{aide}.
DIRE-G denotes ProGAN training and DIRE-D denotes ADM training in the original
comparison. The original tables do not supply complete per-row provenance for
every other baseline. Method coverage differs between accuracy and AP, so a
missing row is not interpreted as a zero score. The evaluation and checkpoint
selection qualifications in Section~\ref{sec:protocol} apply throughout.

\subsection{Unperturbed images}
\input{tables/appendix_aigcd_acc}

\clearpage
\input{tables/appendix_aigcd_ap}

\clearpage
\subsection{JPEG compression}
JPEG affects the models differently across generators. These tables preserve
the complete comparison, with the unresolved quality-factor label noted locally.
\input{tables/appendix_aigcd_acc_jpg}

\input{tables/appendix_aigcd_ap_jpg}

\clearpage
\subsection{Downsampling}
The following results use the study's downsampling setting, with spatial scale
reduced by a factor of 0.5.
\input{tables/appendix_aigcd_acc_down}

\input{tables/appendix_aigcd_ap_down}

\clearpage
\subsection{Gaussian blur}
These results use the study's Gaussian-blur setting with $\sigma=1$.
\input{tables/appendix_aigcd_acc_blur}

\input{tables/appendix_aigcd_ap_blur}

\clearpage
\section{Ablation Details}
\label{app:ablation-details}

\subsection{Supervision and source-image resolution}
The per-generator results separate the behavior of training-domain generators
from transfer to other generators. Image-level supervision and patch-level
supervision both consume patches; they differ in where predictions are combined
relative to the loss. The resolution-aligned variants crop an already resized
source image, complementing this supervision comparison.
\input{tables/appendix_supervision}

\input{tables/appendix_aligned}

\clearpage
\subsection{Patch size}
A change of crop size can help on one generator and hurt on another. The
detailed results below therefore accompany the two benchmark means.
\input{tables/appendix_patch_size}

\clearpage
\subsection{Inference coverage}
Increasing the number of patches chiefly helps between one and a modest
number of crops in this experiment. Beyond 64, changes are small and need not
be positive for each generator. Timing records are included separately because
their measurement scope does not support a standardized latency claim.
\input{tables/appendix_patch_number}

\input{tables/appendix_timing}

\clearpage
\section{Single-Patch Selection and Transfer}
\label{app:single-transfer}

These experiments examine the single-patch setting that motivated the
multi-patch method. SPD selects one RGB patch per image instance; it does not
include the variational bottleneck of EIB-Net~\citep{eib2026}. The single-patch
experiments have their own preprocessing and optimization settings and are
not pooled numerically with the PWS main experiments.

\subsection{Selection, grid size and seed}
The minimum-, middle- and maximum-complexity strategies use the local variation
statistic described in Appendix~\ref{sec:implementation}; random selection does
not use that statistic. The detailed records retain the selection, candidate
grid size, training fraction and seed rather than collapsing them to the best
score. The recorded convergence rule uses a 20-epoch validation moving average,
a deviation within two percentage points, and an absolute moving-average slope
below 0.01.
\input{tables/appendix_single_selection}

\clearpage
\subsection{Cross-scene and cross-generator transfer}
The DiffusionForensics study trains and validates on LSUN-Bedroom and evaluates
on ImageNet-based subsets. It tests a change of image content together with
transfer to SDv1 generation, beyond the ADM validation generator.
\input{tables/appendix_diffusion_transfer}

\clearpage
\subsection{Training-data fraction, mixed sources and model capacity}
Single-patch inputs do not improve every full-data result. Their gains in the
following low-data comparisons and mixed-source setting are useful complements
to the main multi-patch evaluation, without implying that one crop is always
preferable to the whole image.
\input{tables/appendix_diff_intra}

\input{tables/appendix_mixed_source}

\input{tables/appendix_lightweight}

%% file: tables/appendix_aigcd_acc.tex
% Source: aigcd/table/sup/aigcd_acc.tex. Numerical entries retained.
\begin{center}
\begin{minipage}{\textwidth}
\centering
\small
\renewcommand{\arraystretch}{1.12}
\captionof{table}{AIGCDetectBenchmark accuracy (\%) without additional distortion.}
\label{tab:aigcd_acc}
\label{tab:appendix-aigcd}
\textit{(a) ProGAN through WFR}\par\smallskip
\begin{tabular*}{\linewidth}{@{\extracolsep{\fill}}lrrrrrrrr@{}}
\toprule
Method & ProGAN & StyleGAN & BigGAN & CycleGAN & StarGAN & GauGAN & StyleGAN2 & WFR \\
\midrule
CNNSpot & 100.0 & 90.17 & 71.17 & 87.62 & 94.60 & 81.42 & 86.91 & 91.65 \\
LNP & 99.67 & 91.75 & 77.75 & 84.10 & 99.92 & 75.39 & 94.64 & 70.85 \\
LGrad & 99.83 & 91.08 & 85.62 & 86.94 & 99.27 & 78.46 & 85.32 & 55.70 \\
DIRE-G & 95.19 & 83.03 & 70.12 & 74.19 & 95.47 & 67.79 & 75.31 & 58.05 \\
DIRE-D & 52.75$^{\ddagger}$ & 51.31 & 49.70 & 49.58 & 46.72 & 51.23 & 51.72 & 53.30 \\
PatchCraft & 100.0 & 92.77 & 95.80 & 70.17 & 99.97 & 71.58 & 89.55 & 85.80 \\
NPR & 99.79 & 97.70 & 84.35 & 96.10 & 99.35 & 82.50 & 98.38 & 65.80 \\
AIDE & 99.99 & 99.64 & 83.95 & 98.48 & 99.91 & 73.25 & 98.00 & 94.20 \\
\midrule
PWS (ResNet18) & 99.99 & 88.11 & 70.05 & 89.21 & 100.0 & 57.08 & 95.34 & 98.85 \\
PWS (ResNet50) & 99.99 & 88.93 & 76.02 & 90.92 & 99.97 & 62.45 & 94.74 & 87.15 \\
PWS (Xception) & 100.0 & 95.89 & 79.65 & 96.48 & 98.70 & 74.73 & 99.58 & 96.45 \\
PWS (Swin-T) & 100.0 & 98.66 & 83.62 & 98.98 & 100.0 & 76.80 & 99.00 & 80.60 \\
\bottomrule
\end{tabular*}
\par\medskip
\textit{(b) ADM through DALL-E2 and the overall mean}\par\smallskip
\begin{tabular*}{\linewidth}{@{\extracolsep{\fill}}lrrrrrrrrr@{}}
\toprule
Method & ADM & GLIDE & Midj. & SD1.4 & SD1.5 & VQDM & Wukong & DALL-E2 & Mean \\
\midrule
CNNSpot & 60.39 & 58.07 & 51.39 & 50.57 & 50.53 & 56.46 & 51.03 & 50.45 & 70.78 \\
LNP & 84.73 & 80.52 & 65.55 & 85.55 & 85.67 & 74.46 & 82.06 & 88.75 & 83.84 \\
LGrad & 67.15 & 66.11 & 65.35 & 63.02 & 63.67 & 72.99 & 59.55 & 65.45 & 75.34 \\
DIRE-G & 75.78 & 71.75 & 58.01 & 49.74 & 49.83 & 53.68 & 54.46 & 66.48 & 68.68 \\
DIRE-D & 98.25 & 92.42 & 89.45 & 91.24 & 91.63 & 91.90 & 90.90 & 92.45 & 71.53 \\
PatchCraft & 82.17 & 83.79 & 90.12 & 95.38 & 95.30 & 88.91 & 91.07 & 96.60 & 89.31 \\
NPR & 69.69 & 78.36 & 77.85 & 78.63 & 78.89 & 78.13 & 76.11 & 64.90 & 82.91 \\
AIDE & 93.43 & 95.09 & 77.20 & 93.00 & 92.85 & 95.16 & 93.55 & 96.60 & 92.77 \\
\midrule
PWS (ResNet18) & 86.54 & 90.01 & 78.84 & 95.17 & 95.58 & 83.72 & 89.78 & 96.30 & 88.41 \\
PWS (ResNet50) & 88.58 & 89.01 & 85.08 & 94.88 & 94.91 & 90.28 & 95.08 & 91.65 & 89.35 \\
PWS (Xception) & 84.05 & 94.10 & 84.87 & 97.87 & 97.67 & 91.49 & 99.51 & 97.25 & 93.02 \\
PWS (Swin-T) & 95.03 & 90.25 & 84.91 & 98.26 & 98.32 & 92.25 & 99.38 & 98.85 & 93.43 \\
\bottomrule
\end{tabular*}
\par\smallskip\raggedright\footnotesize $\ddagger$ The ProGAN entry is corrected to 52.75 from two other saved copies of the same table; the source appendix copy contains the transcription error ``52..75''. The corrected row reproduces the reported mean of 71.53.
\end{minipage}
\end{center}

%% file: tables/appendix_aigcd_ap.tex
% Source: aigcd/table/sup/aigcd_ap.tex. Numerical entries retained.
\begin{center}
\begin{minipage}{\textwidth}
\centering
\small
\renewcommand{\arraystretch}{1.12}
\captionof{table}{AIGCDetectBenchmark average precision (\%) without additional distortion.}
\label{tab:aigcd_ap}
\textit{(a) ProGAN through WFR}\par\smallskip
\begin{tabular*}{\linewidth}{@{\extracolsep{\fill}}lrrrrrrrr@{}}
\toprule
Method & ProGAN & StyleGAN & BigGAN & CycleGAN & StarGAN & GauGAN & StyleGAN2 & WFR \\
\midrule
CNNSpot & 100.0 & 99.83 & 85.99 & 94.94 & 99.04 & 90.82 & 99.48 & 99.85 \\
LNP & 100.0 & 99.27 & 94.54 & 89.52 & 100.0 & 84.54 & 99.70 & 42.75 \\
LGrad & 100.0 & 98.31 & 92.93 & 95.01 & 100.0 & 95.43 & 97.89 & 57.99 \\
DIRE-G & 99.08 & 91.74 & 75.25 & 80.56 & 99.34 & 72.15 & 88.30 & 60.13 \\
UnivFD & 100.0 & 97.56 & 99.27 & 99.80 & 99.37 & 99.98 & 97.90 & 96.73 \\
PatchCraft & 100.0 & 98.96 & 99.42 & 85.26 & 100.0 & 81.33 & 97.74 & 95.26 \\
\midrule
PWS (ResNet18) & 100.0 & 98.89 & 76.53 & 95.94 & 100.0 & 52.55 & 99.98 & 99.96 \\
PWS (ResNet50) & 100.0 & 99.11 & 83.81 & 96.73 & 100.0 & 64.49 & 99.97 & 99.49 \\
PWS (Xception) & 100.0 & 99.99 & 87.18 & 98.58 & 100.0 & 91.47 & 100.0 & 99.94 \\
PWS (Swin-T) & 100.0 & 100.0 & 91.35 & 99.93 & 100.0 & 92.24 & 100.0 & 99.44 \\
\bottomrule
\end{tabular*}
\par\medskip
\textit{(b) ADM through DALL-E2 and the overall mean}\par\smallskip
\begin{tabular*}{\linewidth}{@{\extracolsep{\fill}}lrrrrrrrrr@{}}
\toprule
Method & ADM & GLIDE & Midj. & SD1.4 & SD1.5 & VQDM & Wukong & DALL-E2 & Mean \\
\midrule
CNNSpot & 75.67 & 72.28 & 66.24 & 61.20 & 61.56 & 68.83 & 57.34 & 53.51 & 80.41 \\
LNP & 93.37 & 92.76 & 86.92 & 96.34 & 96.00 & 95.91 & 95.33 & 98.26 & 91.58 \\
LGrad & 72.95 & 80.42 & 71.86 & 62.37 & 62.85 & 77.47 & 62.48 & 82.55 & 81.91 \\
DIRE-G & 85.84 & 78.35 & 61.86 & 91.86 & 91.52 & 54.57 & 55.38 & 74.48 & 78.78 \\
UnivFD & 86.81 & 83.81 & 74.00 & 86.14 & 85.84 & 96.53 & 91.07 & 63.04 & 91.74$^{\dagger}$ \\
PatchCraft & 93.40 & 94.04 & 96.44 & 99.06 & 99.06 & 96.26 & 97.54 & 99.56 & 95.83 \\
\midrule
PWS (ResNet18) & 95.46 & 93.11 & 98.97 & 99.03 & 98.92 & 94.66 & 99.95 & 99.57 & 93.97 \\
PWS (ResNet50) & 95.65 & 96.08 & 92.64 & 99.28 & 99.08 & 96.52 & 99.87 & 99.02 & 95.11 \\
PWS (Xception) & 95.19 & 98.16 & 95.77 & 99.51 & 99.41 & 97.77 & 100.0 & 99.89 & 97.69$^{\dagger}$ \\
PWS (Swin-T) & 98.83 & 97.63 & 97.72 & 99.72 & 99.67 & 99.17 & 99.99 & 99.92 & 98.48 \\
\bottomrule
\end{tabular*}
\par\smallskip\raggedright\footnotesize $\dagger$ Reported means are retained: the displayed UnivFD and PWS (Xception) entries average to 91.115625 and 97.67875, respectively, rather than 91.74 and 97.69. These discrepancies require checking against the underlying results.
\end{minipage}
\end{center}

%% file: tables/appendix_aigcd_acc_jpg.tex
% Source: aigcd/table/sup/aigcd_acc_jpg.tex. Numerical entries retained.
\begin{center}
\begin{minipage}{\textwidth}
\centering
\small
\renewcommand{\arraystretch}{1.12}
\captionof{table}{AIGCDetectBenchmark accuracy (\%) after JPEG compression.}
\label{tab:aigcd_acc_jpg}
\textit{(a) ProGAN through WFR}\par\smallskip
\begin{tabular*}{\linewidth}{@{\extracolsep{\fill}}lrrrrrrrr@{}}
\toprule
Method & ProGAN & StyleGAN & BigGAN & CycleGAN & StarGAN & GauGAN & StyleGAN2 & WFR \\
\midrule
CNNSpot & 99.96 & 75.00 & 62.05 & 83.16 & 79.26 & 69.89 & 71.29 & 79.10 \\
FreDect & 84.40 & 72.30 & 62.90 & 71.35 & 83.79 & 64.99 & 73.34 & 50.25 \\
Fusing & 99.64 & 70.45 & 63.50 & 80.96 & 92.02 & 66.91 & 60.19 & 59.90 \\
GramNet & 99.94 & 77.74 & 62.86 & 86.98 & 88.72 & 65.69 & 78.87 & 86.40 \\
LNP & 67.80 & 55.42 & 51.73 & 62.34 & 52.00 & 50.18 & 58.32 & 53.15 \\
LGrad & 55.70 & 55.90 & 51.35 & 57.15 & 51.68 & 49.70 & 55.99 & 52.90 \\
DIRE-G & 98.92 & 75.93 & 67.85 & 74.68 & 83.08 & 77.84 & 68.99 & 59.05 \\
DIRE-D & 51.55 & 50.89 & 51.08 & 50.53 & 41.55 & 39.31 & 51.36 & 54.35 \\
UnivFD & 99.34 & 79.74 & 88.22 & 98.61 & 95.32 & 98.61 & 69.41 & 70.20 \\
PatchCraft & 97.84 & 82.49 & 65.25 & 71.99 & 60.21 & 73.71 & 82.71 & 79.40 \\
\midrule
PWS (Swin-T) & 99.84 & 97.06 & 77.35 & 86.98 & 98.82 & 74.47 & 97.03 & 80.75 \\
\bottomrule
\end{tabular*}
\par\medskip
\textit{(b) ADM through DALL-E2 and the overall mean}\par\smallskip
\begin{tabular*}{\linewidth}{@{\extracolsep{\fill}}lrrrrrrrrr@{}}
\toprule
Method & ADM & GLIDE & Midj. & SD1.4 & SD1.5 & VQDM & Wukong & DALL-E2 & Mean \\
\midrule
CNNSpot & 51.28 & 51.92 & 50.90 & 49.82 & 49.90 & 51.24 & 50.01 & 49.70 & 64.03 \\
FreDect & 67.72 & 66.13 & 56.61 & 52.95 & 52.71 & 76.51 & 52.60 & 82.70 & 66.95 \\
Fusing & 50.01 & 51.99 & 50.85 & 50.16 & 50.13 & 51.61 & 50.00 & 50.00 & 62.40 \\
GramNet & 51.62 & 51.59 & 49.85 & 49.44 & 49.56 & 49.36 & 49.62 & 48.20 & 65.40 \\
LNP & 51.22 & 51.28 & 50.14 & 51.84 & 51.87 & 50.98 & 50.91 & 50.70 & 53.74 \\
LGrad & 46.50 & 45.09 & 62.32 & 53.87 & 53.62 & 44.06 & 52.49 & 36.40 & 51.54 \\
DIRE-G & 62.49 & 73.11 & 58.90 & 48.40 & 48.48 & 54.82 & 52.14 & 60.40 & 66.57 \\
DIRE-D & 91.92 & 91.38 & 90.78 & 92.53 & 92.56 & 92.47 & 92.14 & 89.89 & 70.27 \\
UnivFD & 64.41 & 64.14 & 55.43 & 56.48 & 56.10 & 78.89 & 62.75 & 50.35 & 74.25 \\
PatchCraft & 62.64 & 68.01 & 57.87 & 75.00 & 74.87 & 64.94 & 67.91 & 70.35 & 72.20 \\
\midrule
PWS (Swin-T) & 78.97 & 84.27 & 69.12 & 71.47 & 71.21 & 69.94 & 50.10 & 84.00 & 80.71 \\
\bottomrule
\end{tabular*}
\par\smallskip\raggedright\footnotesize The retained tables do not identify the JPEG quality factor unambiguously; these entries are not assigned to QF=90 or QF=95.
\end{minipage}
\end{center}

%% file: tables/appendix_aigcd_ap_jpg.tex
% Source: aigcd/table/sup/aigcd_ap_jpg.tex. Numerical entries retained.
\begin{center}
\begin{minipage}{\textwidth}
\centering
\small
\renewcommand{\arraystretch}{1.12}
\captionof{table}{AIGCDetectBenchmark average precision (\%) after JPEG compression.}
\label{tab:aigcd_ap_jpg}
\textit{(a) ProGAN through WFR}\par\smallskip
\begin{tabular*}{\linewidth}{@{\extracolsep{\fill}}lrrrrrrrr@{}}
\toprule
Method & ProGAN & StyleGAN & BigGAN & CycleGAN & StarGAN & GauGAN & StyleGAN2 & WFR \\
\midrule
CNNSpot & 100.0 & 98.60 & 88.36 & 96.08 & 93.63 & 95.53 & 98.08 & 94.34 \\
GramNet & 100.0 & 98.24 & 83.14 & 96.32 & 97.16 & 84.55 & 98.42 & 95.19 \\
LNP & 79.86 & 79.86 & 67.56 & 73.69 & 76.68 & 60.98 & 87.56 & 69.81 \\
LGrad & 68.65 & 75.48 & 54.21 & 70.26 & 61.44 & 51.42 & 77.57 & 55.02 \\
DIRE-G & 99.94 & 94.12 & 78.07 & 84.29 & 92.16 & 76.22 & 92.02 & 61.86 \\
DIRE-D & 50.95 & 53.59 & 50.21 & 52.13 & 39.44 & 37.73 & 54.50 & 60.86 \\
UnivFD & 99.98 & 94.90 & 97.65 & 99.47 & 99.07 & 99.91 & 94.65 & 88.13 \\
PatchCraft & 99.73 & 91.12 & 69.41 & 83.30 & 70.29 & 82.41 & 91.03 & 89.49 \\
\midrule
PWS (Swin-T) & 100.0 & 99.82 & 82.53 & 95.64 & 99.97 & 76.71 & 99.82 & 90.69 \\
\bottomrule
\end{tabular*}
\par\medskip
\textit{(b) ADM through DALL-E2 and the overall mean}\par\smallskip
\begin{tabular*}{\linewidth}{@{\extracolsep{\fill}}lrrrrrrrrr@{}}
\toprule
Method & ADM & GLIDE & Midj. & SD1.4 & SD1.5 & VQDM & Wukong & DALL-E2 & Mean \\
\midrule
CNNSpot & 64.18 & 69.60 & 61.95 & 55.52 & 56.47 & 64.82 & 54.78 & 47.03 & 77.44 \\
GramNet & 60.64 & 62.67 & 54.79 & 52.56 & 53.32 & 58.96 & 51.40 & 43.52 & 74.43 \\
LNP & 61.90 & 69.27 & 63.94 & 70.53 & 70.51 & 60.51 & 66.00 & 55.01 & 69.60 \\
LGrad & 44.73 & 42.16 & 68.04 & 56.03 & 56.30 & 40.91 & 54.32 & 36.07 & 57.04 \\
DIRE-G & 71.20 & 83.60 & 65.10 & 45.14 & 45.22 & 58.94 & 50.32 & 66.37 & 72.79 \\
DIRE-D & 98.99 & 98.44 & 97.96 & 99.15 & 99.19 & 99.39 & 99.23 & 99.51 & 74.45 \\
UnivFD & 84.00 & 85.18 & 71.41 & 75.23 & 74.82 & 93.58 & 83.69 & 58.95 & 87.54 \\
PatchCraft & 75.07 & 81.04 & 66.42 & 86.47 & 86.59 & 78.54 & 80.26 & 95.59 & 82.92 \\
\midrule
PWS (Swin-T) & 87.86 & 92.85 & 77.41 & 78.71 & 77.87 & 78.44 & 55.18 & 92.92 & 86.65 \\
\bottomrule
\end{tabular*}
\par\smallskip\raggedright\footnotesize The retained tables do not identify the JPEG quality factor unambiguously; these entries are not assigned to QF=90 or QF=95.
\end{minipage}
\end{center}

%% file: tables/appendix_aigcd_acc_down.tex
% Source: aigcd/table/sup/aigcd_acc_down.tex. Numerical entries retained.
\begin{center}
\begin{minipage}{\textwidth}
\centering
\small
\renewcommand{\arraystretch}{1.12}
\captionof{table}{AIGCDetectBenchmark accuracy (\%) after downsampling by a factor of 0.5.}
\label{tab:aigcd_acc_down}
\textit{(a) ProGAN through WFR}\par\smallskip
\begin{tabular*}{\linewidth}{@{\extracolsep{\fill}}lrrrrrrrr@{}}
\toprule
Method & ProGAN & StyleGAN & BigGAN & CycleGAN & StarGAN & GauGAN & StyleGAN2 & WFR \\
\midrule
CNNSpot & 88.00 & 64.46 & 52.02 & 60.83 & 64.58 & 65.61 & 65.69 & 76.50 \\
LNP & 85.15 & 80.34 & 70.95 & 67.79 & 86.44 & 53.89 & 89.16 & 51.90 \\
LGrad & 81.46 & 71.32 & 58.23 & 52.42 & 58.38 & 55.04 & 69.93 & 56.70 \\
DIRE-G & 68.01 & 66.52 & 52.08 & 57.53 & 59.80 & 44.82 & 66.79 & 50.85 \\
DIRE-D & 49.74 & 51.74 & 50.68 & 46.21 & 58.39 & 52.61 & 60.05 & 50.00 \\
UnivFD & 95.83 & 72.63 & 73.00 & 89.21 & 88.02 & 91.23 & 68.66 & 73.75 \\
PatchCraft & 99.92 & 90.37 & 72.35 & 83.76 & 99.90 & 62.07 & 89.00 & 79.55 \\
\midrule
PWS (Swin-T) & 99.36 & 95.99 & 59.00 & 67.64 & 100.0 & 54.20 & 99.34 & 50.05 \\
\bottomrule
\end{tabular*}
\par\medskip
\textit{(b) ADM through DALL-E2 and the overall mean}\par\smallskip
\begin{tabular*}{\linewidth}{@{\extracolsep{\fill}}lrrrrrrrrr@{}}
\toprule
Method & ADM & GLIDE & Midj. & SD1.4 & SD1.5 & VQDM & Wukong & DALL-E2 & Mean \\
\midrule
CNNSpot & 50.09 & 50.21 & 51.92 & 51.42 & 51.66 & 50.24 & 50.31 & 48.00 & 58.85 \\
LNP & 66.25 & 53.83 & 49.06 & 48.28 & 47.81 & 51.86 & 51.15 & 62.90 & 63.55 \\
LGrad & 55.89 & 58.52 & 62.26 & 59.24 & 58.82 & 57.96 & 56.65 & 60.95 & 60.86 \\
DIRE-G & 54.63 & 58.92 & 53.67 & 50.80 & 49.36 & 52.56 & 52.36 & 58.75 & 56.09 \\
DIRE-D & 75.27 & 71.67 & 69.56 & 74.48 & 74.39 & 76.06 & 69.62 & 65.75 & 62.26 \\
UnivFD & 71.94 & 69.56 & 50.40 & 51.17 & 51.04 & 81.45 & 54.64 & 51.40 & 70.87 \\
PatchCraft & 71.12 & 58.37 & 57.87 & 81.39 & 81.01 & 75.30 & 78.74 & 73.40 & 78.38 \\
\midrule
PWS (Swin-T) & 95.12 & 91.74 & 89.03 & 98.60 & 98.56 & 98.60 & 93.08 & 98.95 & 86.83 \\
\bottomrule
\end{tabular*}
\end{minipage}
\end{center}

%% file: tables/appendix_aigcd_ap_down.tex
% Source: aigcd/table/sup/aigcd_ap_down.tex. Numerical entries retained.
\begin{center}
\begin{minipage}{\textwidth}
\centering
\small
\renewcommand{\arraystretch}{1.12}
\captionof{table}{AIGCDetectBenchmark average precision (\%) after downsampling by a factor of 0.5.}
\label{tab:aigcd_ap_down}
\textit{(a) ProGAN through WFR}\par\smallskip
\begin{tabular*}{\linewidth}{@{\extracolsep{\fill}}lrrrrrrrr@{}}
\toprule
Method & ProGAN & StyleGAN & BigGAN & CycleGAN & StarGAN & GauGAN & StyleGAN2 & WFR \\
\midrule
CNNSpot & 99.27 & 90.80 & 61.36 & 77.25 & 93.20 & 85.10 & 91.65 & 84.73 \\
GramNet & 98.94 & 90.25 & 61.15 & 82.52 & 95.17 & 82.65 & 91.46 & 86.02 \\
LNP & 94.12 & 93.49 & 75.77 & 71.89 & 99.50 & 52.67 & 97.11 & 79.90 \\
LGrad & 97.21 & 91.79 & 64.29 & 58.24 & 97.29 & 57.91 & 94.44 & 61.07 \\
DIRE-G & 77.93 & 80.44 & 52.73 & 60.26 & 69.22 & 45.73 & 77.45 & 51.34 \\
DIRE-D & 50.18 & 56.19 & 52.23 & 46.40 & 47.37 & 40.67 & 70.22 & 51.30 \\
UnivFD & 99.35 & 90.72 & 84.65 & 96.34 & 95.72 & 97.51 & 87.12 & 89.00 \\
PatchCraft & 100.0 & 99.17 & 72.34 & 92.49 & 100.0 & 67.57 & 99.17 & 91.22 \\
\midrule
PWS (Swin-T) & 100.0 & 99.84 & 63.44 & 88.21 & 100.0 & 64.34 & 99.98 & 58.38 \\
\bottomrule
\end{tabular*}
\par\medskip
\textit{(b) ADM through DALL-E2 and the overall mean}\par\smallskip
\begin{tabular*}{\linewidth}{@{\extracolsep{\fill}}lrrrrrrrrr@{}}
\toprule
Method & ADM & GLIDE & Midj. & SD1.4 & SD1.5 & VQDM & Wukong & DALL-E2 & Mean \\
\midrule
CNNSpot & 61.52 & 60.63 & 54.47 & 55.50 & 55.97 & 59.25 & 51.57 & 55.21 & 71.09 \\
GramNet & 57.21 & 55.83 & 54.78 & 53.80 & 53.68 & 55.79 & 50.39 & 50.55 & 70.01 \\
LNP & 76.01 & 58.44 & 48.75 & 47.75 & 47.63 & 54.33 & 50.39 & 72.05 & 69.99 \\
LGrad & 60.45 & 65.90 & 67.83 & 62.72 & 63.31 & 57.93 & 52.91 & 72.05 & 70.33 \\
DIRE-G & 60.45 & 71.15 & 58.95 & 51.93 & 50.42 & 57.93 & 52.31 & 69.84 & 61.76 \\
DIRE-D & 99.62 & 99.53 & 99.18 & 99.71 & 99.54 & 99.54 & 99.61 & 99.58 & 75.68 \\
UnivFD & 91.77 & 91.14 & 52.57 & 62.25 & 62.56 & 95.71 & 72.80 & 92.99 & 85.14 \\
PatchCraft & 81.07 & 65.03 & 63.24 & 91.01 & 91.10 & 84.03 & 87.03 & 85.78 & 85.64 \\
\midrule
PWS (Swin-T) & 98.29 & 98.07 & 97.81 & 99.87 & 99.85 & 99.90 & 98.20 & 99.98 & 91.64 \\
\bottomrule
\end{tabular*}
\end{minipage}
\end{center}

%% file: tables/appendix_aigcd_acc_blur.tex
% Source: aigcd/table/sup/aigcd_acc_blur.tex. Numerical entries retained.
\begin{center}
\begin{minipage}{\textwidth}
\centering
\small
\renewcommand{\arraystretch}{1.12}
\captionof{table}{AIGCDetectBenchmark accuracy (\%) after Gaussian blur ($\sigma=1$).}
\label{tab:aigcd_acc_blur}
\textit{(a) ProGAN through WFR}\par\smallskip
\begin{tabular*}{\linewidth}{@{\extracolsep{\fill}}lrrrrrrrr@{}}
\toprule
Method & ProGAN & StyleGAN & BigGAN & CycleGAN & StarGAN & GauGAN & StyleGAN2 & WFR \\
\midrule
CNNSpot & 99.95 & 83.32 & 68.03 & 85.65 & 87.12 & 79.30 & 84.04 & 82.70 \\
GramNet & 99.90 & 84.84 & 67.42 & 86.41 & 92.95 & 72.98 & 87.49 & 82.40 \\
LNP & 76.56 & 68.36 & 63.90 & 52.65 & 63.16 & 49.23 & 76.75 & 70.30 \\
LGrad & 95.46 & 85.26 & 63.90 & 53.48 & 92.25 & 61.09 & 75.06 & 53.35 \\
DIRE-G & 85.85 & 72.79 & 57.35 & 65.44 & 80.55 & 62.72 & 63.20 & 61.15 \\
DIRE-D & 51.84 & 50.75 & 51.28 & 46.40 & 41.52 & 39.23 & 51.04 & 52.80 \\
UnivFD & 98.65 & 71.99 & 76.92 & 94.66 & 89.62 & 97.46 & 62.11 & 58.55 \\
PatchCraft & 99.01 & 90.38 & 63.00 & 75.47 & 78.71 & 60.65 & 91.99 & 62.30 \\
\midrule
PWS (Swin-T) & 99.58 & 97.04 & 54.35 & 67.07 & 100.0 & 55.59 & 99.66 & 49.65 \\
\bottomrule
\end{tabular*}
\par\medskip
\textit{(b) ADM through DALL-E2 and the overall mean}\par\smallskip
\begin{tabular*}{\linewidth}{@{\extracolsep{\fill}}lrrrrrrrrr@{}}
\toprule
Method & ADM & GLIDE & Midj. & SD1.4 & SD1.5 & VQDM & Wukong & DALL-E2 & Mean \\
\midrule
CNNSpot & 59.30 & 55.28 & 51.37 & 51.23 & 51.46 & 55.56 & 50.62 & 49.30 & 68.39 \\
GramNet & 87.88 & 54.45 & 50.68 & 52.87 & 52.94 & 57.25 & 51.43 & 49.10 & 70.69 \\
LNP & 70.39 & 83.31 & 68.52 & 64.31 & 64.11 & 57.05 & 60.68 & 85.90 & 67.20 \\
LGrad & 70.99 & 78.54 & 77.43 & 62.93 & 67.32 & 63.28 & 60.09 & 80.20 & 71.29 \\
DIRE-G & 93.13 & 75.14 & 55.44 & 47.04 & 47.21 & 93.43 & 51.63 & 90.60 & 68.92 \\
DIRE-D & 93.13 & 92.44 & 90.42 & 92.87 & 92.54 & 92.87 & 92.83 & 99.05 & 70.69 \\
UnivFD & 64.50 & 60.88 & 55.48 & 54.56 & 54.20 & 76.47 & 58.59 & 92.99 & 72.98 \\
PatchCraft & 69.58 & 72.52 & 76.28 & 78.85 & 78.61 & 70.53 & 74.23 & 72.00 & 75.88 \\
\midrule
PWS (Swin-T) & 94.53 & 95.92 & 93.44 & 97.39 & 97.46 & 95.97 & 98.75 & 98.45 & 87.18 \\
\bottomrule
\end{tabular*}
\end{minipage}
\end{center}

%% file: tables/appendix_aigcd_ap_blur.tex
% Source: aigcd/table/sup/aigcd_ap_blur.tex. Numerical entries retained.
\begin{center}
\begin{minipage}{\textwidth}
\centering
\small
\renewcommand{\arraystretch}{1.12}
\captionof{table}{AIGCDetectBenchmark average precision (\%) after Gaussian blur ($\sigma=1$).}
\label{tab:aigcd_ap_blur}
\textit{(a) ProGAN through WFR}\par\smallskip
\begin{tabular*}{\linewidth}{@{\extracolsep{\fill}}lrrrrrrrr@{}}
\toprule
Method & ProGAN & StyleGAN & BigGAN & CycleGAN & StarGAN & GauGAN & StyleGAN2 & WFR \\
\midrule
CNNSpot & 100.0 & 98.97 & 79.50 & 91.80 & 97.92 & 87.61 & 99.00 & 87.87 \\
GramNet & 100.0 & 98.70 & 79.82 & 94.84 & 98.81 & 84.55 & 99.01 & 91.14 \\
LNP & 87.11 & 79.52 & 54.31 & 52.62 & 88.55 & 47.07 & 88.05 & 49.75 \\
LGrad & 99.20 & 96.50 & 63.90 & 57.65 & 99.78 & 63.91 & 97.35 & 57.24 \\
DIRE-G & 95.01 & 84.15 & 59.17 & 71.58 & 82.09 & 55.99 & 77.42 & 61.85 \\
DIRE-D & 51.10 & 52.58 & 51.23 & 51.60 & 42.96 & 39.11 & 54.18 & 54.18 \\
UnivFD & 99.92 & 92.84 & 91.98 & 98.95 & 96.00 & 99.67 & 90.33 & 68.65 \\
PatchCraft & 99.98 & 97.59 & 64.41 & 80.92 & 98.21 & 65.98 & 98.08 & 75.42 \\
\midrule
PWS (Swin-T) & 99.99 & 99.77 & 62.85 & 80.94 & 100.0 & 69.30 & 99.99 & 58.97 \\
\bottomrule
\end{tabular*}
\par\medskip
\textit{(b) ADM through DALL-E2 and the overall mean}\par\smallskip
\begin{tabular*}{\linewidth}{@{\extracolsep{\fill}}lrrrrrrrrr@{}}
\toprule
Method & ADM & GLIDE & Midj. & SD1.4 & SD1.5 & VQDM & Wukong & DALL-E2 & Mean \\
\midrule
CNNSpot & 69.64 & 63.81 & 55.45 & 56.58 & 56.90 & 64.46 & 54.09 & 49.84 & 75.84 \\
GramNet & 68.40 & 62.91 & 55.03 & 58.99 & 59.65 & 55.64 & 55.64 & 47.65 & 75.67 \\
LNP & 49.35 & 92.53 & 75.00 & 68.81 & 68.11 & 62.44 & 60.67 & 95.03 & 69.93 \\
LGrad & 80.35 & 86.81 & 86.39 & 71.72 & 72.94 & 63.56 & 67.87 & 88.19 & 78.33 \\
DIRE-G & 80.35 & 80.90 & 58.39 & 46.01 & 46.05 & 59.97 & 52.51 & 81.81 & 68.33 \\
DIRE-D & 99.66 & 98.98 & 97.62 & 99.39 & 99.51 & 99.87 & 99.47 & 99.60 & 74.44 \\
UnivFD & 77.25 & 77.25 & 65.53 & 67.54 & 67.24 & 90.23 & 74.59 & 54.28 & 82.02 \\
PatchCraft & 78.88 & 83.25 & 58.93 & 90.80 & 90.65 & 90.23 & 83.75 & 90.81 & 84.24 \\
\midrule
PWS (Swin-T) & 98.58 & 99.27 & 98.83 & 99.65 & 99.59 & 99.23 & 99.97 & 99.91 & 91.35$^{\dagger}$ \\
\bottomrule
\end{tabular*}
\par\smallskip\raggedright\footnotesize $\dagger$ The reported PWS mean is 91.35, whereas the 16 displayed values average to 91.6775. Both the source values and the reported mean are retained pending reconciliation.
\end{minipage}
\end{center}

%% file: tables/appendix_supervision.tex
% Source: aigcd/table/sup/gen_image_patch.tex. Numerical entries retained.
\begin{center}
\begin{minipage}{\textwidth}
\centering
\small
\renewcommand{\arraystretch}{1.12}
\captionof{table}{Per-generator GenImage accuracy (\%) with image-level and patch-level supervision. All variants are trained on SD~v1.4. T marks the documented target-selected result; other rows have incompletely recovered selection histories (Section~\ref{sec:selection}).}
\label{tab:patch_level_sup}
\begin{tabular*}{\linewidth}{@{\extracolsep{\fill}}lrrrrrrrrr@{}}
\toprule
Setting & Midj. & SD1.4 & SD1.5 & ADM & GLIDE & Wukong & VQDM & BigGAN & Mean \\
\midrule
Joint (ResNet50) & 54.08 & 99.92 & 99.88 & 53.33 & 55.00 & 98.33 & 54.25 & 62.83 & 72.40 \\
\midrule
PWS (ResNet50)$^{\mathrm{T}}$ & 95.61 & 99.91 & 99.91 & 88.74 & 98.47 & 99.87 & 94.82 & 85.88 & 95.40 \\
Joint (Xception) & 72.33 & 99.92 & 99.94 & 59.92 & 86.17 & 99.42 & 66.67 & 57.17 & 80.19 \\
\midrule
PWS (Xception) & 84.34 & 99.98 & 99.92 & 83.88 & 95.73 & 99.98 & 99.42 & 98.38 & 95.21 \\
Joint (Swin-T) & 66.92 & 99.83 & 99.88 & 59.58 & 58.92 & 100.0 & 72.67 & 63.83 & 77.70 \\
\midrule
PWS (Swin-T) & 87.62 & 99.95 & 99.88 & 94.93 & 78.08 & 99.93 & 98.52 & 89.69 & 93.58 \\
\bottomrule
\end{tabular*}
\par\smallskip\raggedright\footnotesize Joint averages patch logits within an image before computing a loss; PWS applies a loss to each patch before reduction.
\end{minipage}
\end{center}

%% file: tables/appendix_aligned.tex
% Source: aigcd/table/sup/gen_align.tex. Numerical entries retained.
\begin{center}
\begin{minipage}{\textwidth}
\centering
\small
\renewcommand{\arraystretch}{1.12}
\captionof{table}{Per-generator GenImage accuracy (\%) for the resolution-aligned control. Aligned uses a $224\times224$ source canvas for ResNet50 and a $299\times299$ canvas for Xception before patch extraction. T marks the documented target-selected result; other rows have incompletely recovered selection histories (Section~\ref{sec:selection}).}
\label{tab:gen_align}
\begin{tabular*}{\linewidth}{@{\extracolsep{\fill}}lrrrrrrrrr@{}}
\toprule
Setting & Midj. & SD1.4 & SD1.5 & ADM & GLIDE & Wukong & VQDM & BigGAN & Mean \\
\midrule
ResNet50 / whole & 61.25 & 99.92 & 99.81 & 57.83 & 69.58 & 98.67 & 59.67 & 53.75 & 75.06 \\
ResNet50 / aligned & 85.42 & 99.91 & 99.90 & 70.42 & 87.92 & 99.87 & 62.67 & 75.58 & 85.21 \\
ResNet50 / PWS$^{\mathrm{T}}$ & 95.61 & 99.91 & 99.91 & 88.74 & 98.47 & 99.87 & 94.82 & 85.88 & 95.40 \\
Xception / whole & 60.58 & 99.92 & 99.62 & 65.75 & 84.92 & 98.58 & 67.67 & 58.33 & 79.42 \\
Xception / aligned & 79.38 & 99.90 & 99.84 & 70.05 & 89.38 & 99.61 & 68.38 & 76.91 & 85.43 \\
Xception / PWS & 84.34 & 99.98 & 99.92 & 83.88 & 95.73 & 99.98 & 99.42 & 98.38 & 95.21 \\
\bottomrule
\end{tabular*}
\end{minipage}
\end{center}

%% file: tables/appendix_patch_size.tex
% Source: aigcd/table/sup/patch_size.tex. Numerical entries retained.
\begin{center}
\begin{minipage}{\textwidth}
\centering
\small
\renewcommand{\arraystretch}{1.12}
\captionof{table}{Per-generator accuracy (\%) for Xception with different patch sizes. The first panel reports GenImage and the remaining panels report AIGCDetectBenchmark.}
\label{tab:patch_size_sup}
\textit{GenImage}\par\smallskip
\begin{tabular*}{\linewidth}{@{\extracolsep{\fill}}lrrrrrrrrr@{}}
\toprule
Patch size & Midj. & SD1.4 & SD1.5 & ADM & GLIDE & Wukong & VQDM & BigGAN & Mean \\
\midrule
32$\times$32 & 93.40 & 99.95 & 99.88 & 82.35 & 99.42 & 99.77 & 87.73 & 91.90 & 94.30 \\
64$\times$64 & 84.34 & 99.98 & 99.92 & 83.88 & 95.73 & 99.98 & 99.42 & 98.38 & 95.21 \\
128$\times$128 & 83.57 & 99.92 & 99.96 & 72.45 & 98.80 & 99.94 & 88.37 & 88.61 & 91.45 \\
\bottomrule
\end{tabular*}
\par\medskip
\textit{AIGCDetectBenchmark}\par\smallskip
\textit{(a) ProGAN through WFR}\par\smallskip
\begin{tabular*}{\linewidth}{@{\extracolsep{\fill}}lrrrrrrrr@{}}
\toprule
Patch size & ProGAN & StyleGAN & BigGAN & CycleGAN & StarGAN & GauGAN & StyleGAN2 & WFR \\
\midrule
32$\times$32 & 100.00 & 96.33 & 84.00 & 94.72 & 99.00 & 72.30 & 99.62 & 85.50 \\
64$\times$64 & 100.00 & 95.89 & 79.65 & 96.48 & 98.70 & 74.73 & 99.58 & 96.45 \\
128$\times$128 & 99.88 & 87.99 & 80.00 & 93.21 & 100.00 & 66.10 & 99.75 & 99.50 \\
\bottomrule
\end{tabular*}
\par\medskip
\textit{(b) ADM through DALL-E2 and the overall mean}\par\smallskip
\begin{tabular*}{\linewidth}{@{\extracolsep{\fill}}lrrrrrrrrr@{}}
\toprule
Patch size & ADM & GLIDE & Midj. & SD1.4 & SD1.5 & VQDM & Wukong & DALL-E2 & Mean \\
\midrule
32$\times$32 & 88.42 & 93.50 & 78.83 & 97.67 & 97.50 & 91.25 & 93.25 & 96.50 & 91.77 \\
64$\times$64 & 84.05 & 94.10 & 84.87 & 97.87 & 97.67 & 91.49 & 99.51 & 97.25 & 93.02 \\
128$\times$128 & 90.33 & 91.00 & 88.17 & 96.25 & 97.31 & 86.92 & 93.17 & 95.00 & 91.54 \\
\bottomrule
\end{tabular*}
\par\smallskip\raggedright\footnotesize The 128-pixel means are 91.45 on GenImage and 91.54 on AIGCDetectBenchmark. This follows the detailed entries and corrects their interchange in the earlier summary table.
\end{minipage}
\end{center}

%% file: tables/appendix_patch_number.tex
% Source: aigcd/table/sup/patch_nums.tex. Numerical entries retained.
\begin{center}
\begin{minipage}{\textwidth}
\centering
\small
\renewcommand{\arraystretch}{1.12}
\captionof{table}{Per-generator AIGCDetectBenchmark accuracy (\%) as the number of inference patches varies, using Xception.}
\label{tab:patch_num_sup}
\textit{(a) ProGAN through WFR}\par\smallskip
\begin{tabular*}{\linewidth}{@{\extracolsep{\fill}}lrrrrrrrr@{}}
\toprule
Patches & ProGAN & StyleGAN & BigGAN & CycleGAN & StarGAN & GauGAN & StyleGAN2 & WFR \\
\midrule
1 & 99.83 & 91.44 & 73.25 & 89.44 & 84.82 & 71.03 & 93.31 & 92.95 \\
4 & 99.99 & 95.74 & 78.92 & 95.12 & 97.12 & 75.05 & 99.03 & 92.95 \\
16 & 100.00 & 95.23 & 78.33 & 96.21 & 97.97 & 73.86 & 99.42 & 96.05 \\
64 & 100.00 & 95.89 & 79.65 & 96.48 & 98.70 & 74.73 & 99.58 & 96.45 \\
128 & 100.00 & 95.42 & 78.30 & 96.52 & 98.65 & 73.95 & 99.57 & 96.40 \\
256 & 100.00 & 95.42 & 78.30 & 96.52 & 98.65 & 73.95 & 99.57 & 96.65 \\
\bottomrule
\end{tabular*}
\par\medskip
\textit{(b) ADM through DALL-E2 and the overall mean}\par\smallskip
\begin{tabular*}{\linewidth}{@{\extracolsep{\fill}}lrrrrrrrrr@{}}
\toprule
Patches & ADM & GLIDE & Midj. & SD1.4 & SD1.5 & VQDM & Wukong & DALL-E2 & Mean \\
\midrule
1 & 76.32 & 89.78 & 79.71 & 92.38 & 92.59 & 84.27 & 94.77 & 91.45 & 87.33 \\
4 & 82.28 & 93.14 & 81.90 & 96.52 & 96.39 & 89.75 & 98.18 & 95.90 & 91.75 \\
16 & 82.97 & 93.48 & 83.34 & 97.46 & 97.21 & 90.98 & 99.00 & 97.60 & 92.45 \\
64 & 84.05 & 94.10 & 84.87 & 97.87 & 97.67 & 91.49 & 99.51 & 97.25 & 93.02 \\
128 & 83.05 & 93.58 & 83.84 & 97.76 & 97.58 & 91.41 & 99.40 & 97.35 & 92.67 \\
256 & 83.08 & 93.57 & 83.72 & 97.87 & 97.64 & 91.42 & 99.44 & 97.35 & 92.70 \\
\bottomrule
\end{tabular*}
\end{minipage}
\end{center}

%% file: tables/appendix_timing.tex
% Source: aigcd/table/patch_num.tex. Numerical entries retained.
\begin{center}
\begin{minipage}{\textwidth}
\centering
\small
\renewcommand{\arraystretch}{1.12}
\captionof{table}{Timing records accompanying the inference-patch-count experiment.}
\label{tab:appendix-timing}
\begin{tabular*}{\linewidth}{@{\extracolsep{\fill}}lrrr@{}}
\toprule
Patches & Recorded time (s) & WFR accuracy (\%) & Mean accuracy (\%) \\
\midrule
1 & 6.04 & 92.95 & 87.33 \\
4 & 6.19 & 92.95 & 91.75 \\
16 & 6.69 & 96.05 & 92.45 \\
64 & 9.06 & 96.45 & 93.02 \\
128 & 16.73 & 96.40 & 92.67 \\
256 & 30.08 & 96.65 & 92.70 \\
\bottomrule
\end{tabular*}
\par\smallskip\raggedright\footnotesize The measurement unit of work, batching and timing boundaries are unspecified in the saved table. These values are preserved as accompanying records, not interpreted as per-image latency or a standardized speed comparison.
\end{minipage}
\end{center}

%% file: tables/appendix_single_selection.tex
% Source: aigcd/table/sup/sing_patch.tex. Numerical entries retained.
\begin{center}
\begin{minipage}{\textwidth}
\centering
\small
\renewcommand{\arraystretch}{1.12}
\captionof{table}{Single-patch selection and grid-size study on DIFF with ResNet50. Each instance uses one selected patch; $G^2$ is the number of candidate grid cells, not the number of jointly processed inputs. Accuracies are percentages; ``all'' denotes the full training subset.}
\label{table:patch_performance}
\begin{tabular*}{\linewidth}{@{\extracolsep{\fill}}lrrlrrrrr@{}}
\toprule
Selection & $G^2$ & Seed & Data & Conv. epoch & Conv. acc. & Best epoch & Best acc. & Last acc. \\
\midrule
max & 16 & 3407 & all & 163 & 90.05 & 96 & 94.91 & 90.55 \\
max & 16 & 3407 & 20\% & 96 & 78.31 & 25 & 89.94 & 80.68 \\
max & 64 & 3407 & all & 132 & 93.08 & 119 & 94.26 & 91.93 \\
max & 64 & 3407 & 20\% & 61 & 76.32 & 57 & 83.78 & 79.65 \\
mid & 4 & 3407 & all & 36 & 95.94 & 141 & 99.66 & 98.36 \\
mid & 4 & 3407 & 5\% & - & - & 122 & 89.56 & 86.95 \\
mid & 16 & 3407 & all & 54 & 95.45 & 133 & 97.93 & 96.98 \\
mid & 16 & 3407 & 5\% & 89 & 82.82 & 84 & 86.99 & 84.32 \\
mid & 64 & 3407 & all & 96 & 93.92 & 182 & 96.94 & 95.79 \\
mid & 64 & 3407 & 5\% & 96 & 80.80 & 72 & 86.11 & 84.97 \\
min & 1 & 3407 & all & 160 & 94.72 & 97 & 98.13 & 93.57 \\
min & 4 & 3407 & all & 61 & 93.53 & 146 & 97.86 & 95.83 \\
min & 4 & 3407 & all & - & - & 150 & 90.17 & 88.37 \\
min & 16 & 3407 & all & 30 & 97.09 & 142 & 98.66 & 98.39 \\
min & 16 & 3407 & 5\% & 50 & 86.99 & 84 & 92.00 & 91.51 \\
min & 64 & 3407 & all & 60 & 94.84 & 87 & 97.67 & 96.71 \\
min & 256 & 3407 & all & 90 & 85.88 & 18 & 92.65 & 86.80 \\
ori & - & 3407 & all & 166 & 97.25 & 116 & 99.01 & 97.90 \\
ori & - & 3407 & 20\% & - & - & 116 & 90.90 & 83.32 \\
ori & - & 3407 & 5\% & 173 & 77.74 & 41 & 88.10 & 80.57 \\
random & 16 & 3407 & all & 41 & 94.11 & 165 & 98.78 & 98.36 \\
random & 16 & 3407 & 5\% & 159 & 93.23 & 89 & 94.11 & 92.58 \\
random & 4 & 3407 & 5\% & - & - & 133 & 90.97 & 86.99 \\
random & 64 & 3407 & all & 90 & 96.67 & 132 & 98.24 & 96.02 \\
random & 64 & 3407 & 5\% & 37 & 81.06 & 159 & 93.96 & 91.81 \\
random & 16 & 0 & all & 54 & 96.33 & 67 & 98.78 & 98.01 \\
random & 64 & 0 & all & 36 & 94.41 & 147 & 98.05 & 96.06 \\
random & 16 & 24 & all & 41 & 95.14 & 86 & 98.62 & 97.67 \\
random & 64 & 24 & all & 59 & 96.48 & 172 & 98.36 & 96.79 \\
random & 16 & 100 & all & 31 & 97.05 & 157 & 98.70 & 98.36 \\
random & 16 & 200 & all & 42 & 97.21 & 146 & 98.55 & 98.43 \\
\bottomrule
\end{tabular*}
\par\smallskip\raggedright\footnotesize In the saved data manifests, \texttt{subN} denotes every $N$th line, not $N$ percent. The DIFF training list has 29,602 entries; \texttt{sub5} has 5,921 (approximately 20\%) and \texttt{sub20} has 1,481 (approximately 5\%). No run-to-manifest association has been recovered for the conflicting rows, so filenames alone do not resolve their labels.
\par\smallskip\raggedright\footnotesize Conv. is the recorded validation-plateau epoch; best is the maximum recorded accuracy and last refers to epoch 200. A dash denotes an unreported convergence value. The earlier summary labels the reduced-data values 86.99, 92.00, 88.10 and 94.11 as 20\%, whereas this detailed table labels them 5\%; those entries therefore do not support a matched-data-size comparison until the subset labels are reconciled. Both min/$G^2=4$/seed-3407/full-data rows are retained; their distinguishing configuration is not specified.
\end{minipage}
\end{center}

%% file: tables/appendix_diffusion_transfer.tex
% Source: aigcd/table/sup/DIFFusionF.tex. Numerical entries retained.
\begin{center}
\begin{minipage}{\textwidth}
\centering
\small
\renewcommand{\arraystretch}{1.12}
\captionof{table}{DiffusionForensics transfer accuracy (\%). Training and validation use LSUN-Bedroom; validation uses ADM. The two test subsets use ADM and SDv1 generation on ImageNet. Training fractions refer to the LSUN training set.}
\label{tab:appendix-diffusionforensics}
\label{tab:DIFF_SUP}
\begin{tabular*}{\linewidth}{@{\extracolsep{\fill}}lllrrr@{}}
\toprule
Backbone & Training fraction & Input & Validation ADM & Test ADM & Test SDv1 \\
\midrule
ResNet50 & 20\% & Whole & 100 & 64.73 & 46.18 \\
ResNet50 & 20\% & SPD & 100 & 87.05 & 83.54 \\
ResNet50 & 5\% & Whole & 99.45 & 71.29 & 43.27 \\
ResNet50 & 5\% & SPD & 99.95 & 87.11 & 83.86 \\
ResNet50 & 1\% & Whole & 91.80 & 70.90 & 39.98 \\
ResNet50 & 1\% & SPD & 98.40 & 84.16 & 81.71 \\
RMS-S & 20\% & Whole & 100 & 72.17 & 56.55 \\
RMS-S & 20\% & SPD & 100 & 83.82 & 76.39 \\
RMS-S & 5\% & Whole & 99.85 & 81.20 & 57.67 \\
RMS-S & 5\% & SPD & 100 & 81.37 & 74.45 \\
RMS-S & 1\% & Whole & 98.75 & 72.29 & 53.91 \\
RMS-S & 1\% & SPD & 99.85 & 84.43 & 75.28 \\
ViT-B16 & 20\% & Whole & 96.10 & 60.26 & 34.03 \\
ViT-B16 & 20\% & SPD & 100 & 84.27 & 77.94 \\
ViT-B16 & 5\% & Whole & 92.10 & 57.31 & 34.91 \\
ViT-B16 & 5\% & SPD & 99.80 & 77.90 & 66.98 \\
ViT-B16 & 1\% & Whole & 84.15 & 57.67 & 34.70 \\
ViT-B16 & 1\% & SPD & 88.05 & 57.50 & 44.76 \\
Xception & 20\% & Whole & 92.65 & 74.54 & 52.97 \\
Xception & 20\% & SPD & 98.90 & 83.87 & 74.93 \\
Xception & 5\% & Whole & 96.45 & 76.55 & 52.88 \\
Xception & 5\% & SPD & 98.80 & 78.70 & 66.43 \\
Xception & 1\% & Whole & 90.55 & 68.40 & 38.87 \\
Xception & 1\% & SPD & 98.60 & 83.27 & 74.63 \\
\bottomrule
\end{tabular*}
\par\smallskip\raggedright\footnotesize RMS-S is retained as the original table label; the surrounding study uses the name RMT-S. The stated selection criterion was the first epoch reaching 100\% validation accuracy, but some rows do not attain it and their fallback selection rule is unspecified.
\end{minipage}
\end{center}

%% file: tables/appendix_diff_intra.tex
% Source: aigcd/table/DIFF.tex. Numerical entries retained.
\begin{center}
\begin{minipage}{\textwidth}
\centering
\small
\renewcommand{\arraystretch}{1.12}
\captionof{table}{DIFF intra-dataset accuracy (\%) at four training-data fractions. Whole uses the image; SPD uses one selected patch.}
\label{tab:DIFF}
\begin{tabular*}{\linewidth}{@{\extracolsep{\fill}}llrrrr@{}}
\toprule
Backbone & Input & 100\% & 20\% & 5\% & 2\% \\
\midrule
ResNet50 & Whole & 99.01 & 90.90 & 88.10 & 85.73 \\
ResNet50 & SPD & 97.09 & 95.94 & 92.00 & 86.11 \\
RMT-S & Whole & 98.81 & 93.27 & 84.66 & 70.66 \\
RMT-S & SPD & 99.54 & 98.39 & 92.81 & 87.61 \\
ViT-B16 & Whole & 97.02 & 86.53 & 71.46 & 62.09 \\
ViT-B16 & SPD & 95.07 & 92.16 & 85.92 & 70.73 \\
Xception & Whole & 99.85 & 98.09 & 76.28 & 70.12 \\
Xception & SPD & 98.89 & 97.74 & 94.45 & 92.54 \\
\bottomrule
\end{tabular*}
\end{minipage}
\end{center}

%% file: tables/appendix_mixed_source.tex
% Source: aigcd/table/GenImage_single.tex. Numerical entries retained.
\begin{center}
\begin{minipage}{\textwidth}
\centering
\small
\renewcommand{\arraystretch}{1.12}
\captionof{table}{GenImage mixed-source accuracy (\%). Large uses 1\% of the training data from each generator; Small uses 20\% of that subset (0.2\% of each source).}
\label{tab:GenImage_single}
\begin{tabular*}{\linewidth}{@{\extracolsep{\fill}}llrrrr@{}}
\toprule
Data scale & Input & ResNet50 & RMT-S & ViT-B16 & Xception \\
\midrule
Large & Whole & 95.35 & 98.22 & 91.43 & 91.08 \\
Large & SPD & 96.21 & 98.73 & 95.14 & 94.13 \\
Small & Whole & 87.37 & 94.61 & 83.19 & 85.85 \\
Small & SPD & 89.18 & 95.47 & 86.18 & 88.46 \\
\bottomrule
\end{tabular*}
\end{minipage}
\end{center}

%% file: tables/appendix_lightweight.tex
% Source: aigcd/table/small_network.tex. Numerical entries retained.
\begin{center}
\begin{minipage}{\textwidth}
\centering
\small
\renewcommand{\arraystretch}{1.12}
\captionof{table}{DIFF intra-dataset accuracy (\%) for the lightweight-network study.}
\label{tab:small-networks}
\begin{tabular*}{\linewidth}{@{\extracolsep{\fill}}llrrrr@{}}
\toprule
Backbone & Input & 100\% & 20\% & 5\% & 2\% \\
\midrule
ResNet50 & Whole & 99.46 & 90.90 & 88.10 & 85.73 \\
ResNet50 & SPD & 97.09 & 95.94 & 92.00 & 86.11 \\
ResNet18 & Whole & 98.66 & 92.31 & 85.20 & 82.82 \\
ResNet18 & SPD & 98.81 & 97.09 & 93.04 & 87.26 \\
\bottomrule
\end{tabular*}
\par\smallskip\raggedright\footnotesize The saved lightweight-network comparison reports 99.46 for the full-data ResNet50 whole-image control, whereas Table~\ref{tab:DIFF} reports 99.01. These separate recorded values are preserved; their run-level correspondence is unresolved.
\end{minipage}
\end{center}

%% file: sections/exploratory.tex
\section{Single-Patch Transfer Dynamics}
\label{app:transfer-dynamics}

\begin{figure}[htbp]
\centering
\includegraphics[width=0.82\linewidth]{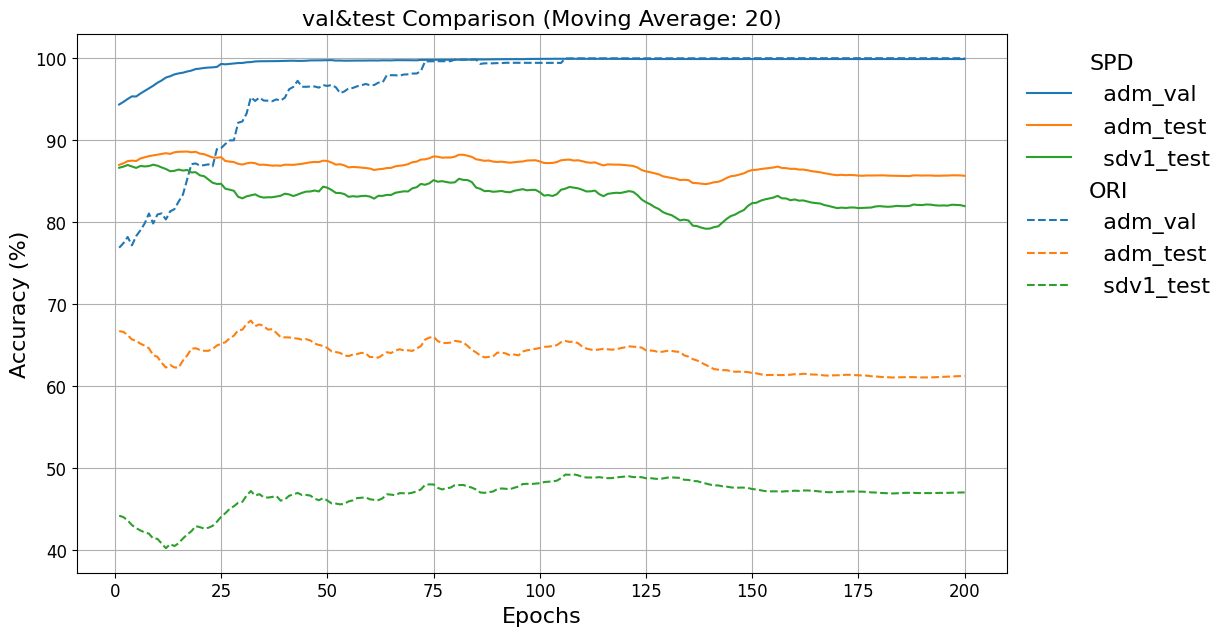}
\caption{Single-patch transfer dynamics with ResNet-50 on DiffusionForensics. Solid lines denote SPD and dashed lines the whole-image baseline (ORI). Blue is ADM validation in the LSUN-Bedroom domain; orange and green are ADM and SD~v1 test results in the ImageNet domain. The historical plot uses a 20-epoch moving average. It illustrates a gap between source validation and target-domain performance, not a confidence interval or a multi-seed estimate.}
\label{fig:single-transfer-curve}
\end{figure}

The single-patch experiments also examine whether high validation accuracy transfers across scene domains. In the DiffusionForensics setting~\citep{dire}, training and validation use the LSUN-Bedroom domain, while the reported test subsets use ImageNet scenes generated by ADM or SD~v1. Figure~\ref{fig:single-transfer-curve} shows a ResNet-50 comparison using the 20\% training-data setting. Both input choices approach saturated ADM validation accuracy, but their target-domain behavior differs: SPD remains substantially above the whole-image baseline on both target curves. The ADM and SD~v1 test curves also differ from one another, despite similarly high source validation accuracy.

This result motivates evaluating a detector beyond its training scene distribution. It supports the usefulness of local input in this particular transfer setting, but does not identify which feature caused the improvement. SPD changes the input representation and its transforms, and these single-patch runs are separate from the multi-patch benchmark study. The smoothed trajectories describe optimization behavior rather than independent repeated trials.

\section{Exploratory Analysis of Local Gradient Statistics}
\label{appendix:gcm}
\label{app:exploratory}

The patch-based formulation is motivated partly by the possibility that nearby-pixel relationships carry evidence beyond convincing global image content. To examine this possibility with a simple descriptor, we analyze a gradient co-occurrence matrix (GCM), adapted from gray-level co-occurrence statistics~\citep{haralick1973textural}. This is a diagnostic analysis only: GCM is not an input, preprocessing step, loss term, or inference module of SPD or PWS.

\subsection{Descriptor and quantization}
The diagnostic scripts convert images to grayscale, resize them to $256\times256$, and compute horizontal and vertical Sobel gradients with a $3\times3$ kernel. Let $M(u)=\sqrt{G_x(u)^2+G_y(u)^2}$ be the magnitude at location $u$, and let $q(M(u))\in\{0,\ldots,15\}$ be its quantized value. With offsets $\mathcal{D}=\{(0,1),(1,0),(1,1),(1,-1)\}$, the normalized co-occurrence matrix is
\begin{equation}
 C_{ab}=\frac{1}{Z}\sum_{\delta\in\mathcal{D}}\sum_{u:\,u+\delta\in\Omega}
 \mathbf{1}\{q(M(u))=a,\ q(M(u+\delta))=b\},
 \qquad H(C)=-\sum_{a,b}C_{ab}\log_2 C_{ab},
\end{equation}
where $Z$ is the number of valid neighbor pairs, $\Omega$ is the image domain, and $0\log 0=0$. Thus, the measured relationships are local, but the statistics are pooled over each resized image; they are not descriptors computed on the detector's sampled $64\times64$ training patches.

The initial analysis uses a linear-range quantization into 16 bins. A second, stratified quantizer redistributes these bins according to each image's median $Q_{50}$, 90th percentile $Q_{90}$, and maximum magnitude $M_{\max}$:
\begin{equation}
 q(m)=\begin{cases}
 \left\lfloor 7.99m/Q_{50}\right\rfloor,&0\leq m<Q_{50},\\[2pt]
 8+\left\lfloor 3.99(m-Q_{50})/(Q_{90}-Q_{50})\right\rfloor,&Q_{50}\leq m<Q_{90},\\[2pt]
 12+\left\lfloor 3.99(m-Q_{90})/(M_{\max}-Q_{90})\right\rfloor,&Q_{90}\leq m\leq M_{\max}.
 \end{cases}
\end{equation}
The implementation separates coincident percentile boundaries by a small positive value before division and clips the result to $[0,15]$. This assigns bins 0--7 to the lower half of the magnitude distribution and four bins to each remaining range. Quantization is part of the diagnostic and can itself change the visible structure. In particular, concentrations around bins 8 and 12 need not be generator-specific signatures.

\subsection{Observed distribution differences}
\begin{figure}[htbp]
\centering
\includegraphics[width=0.88\linewidth]{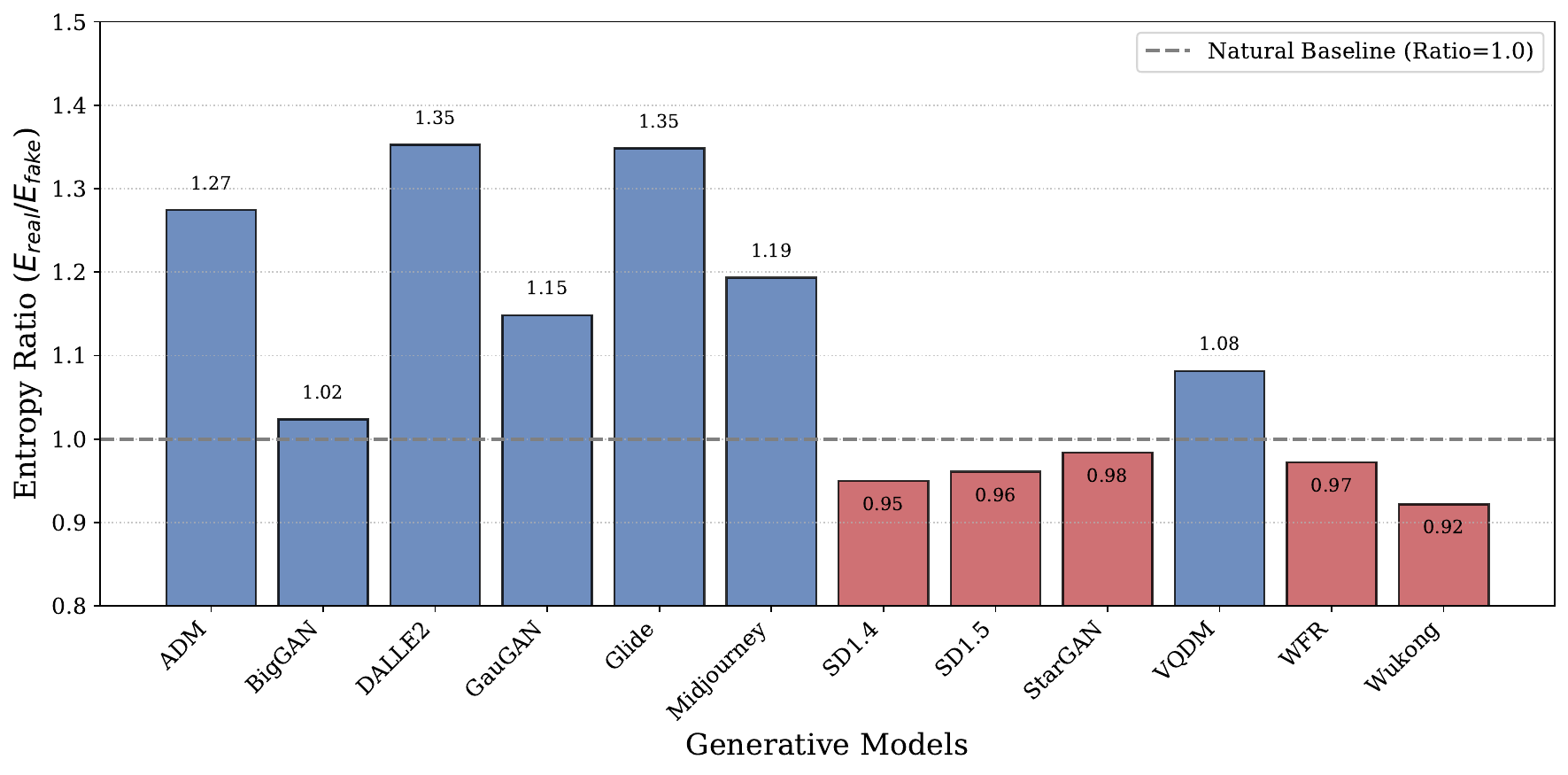}
\caption{Exploratory GCM entropy ratios across 12 synthetic-image subsets. The dashed reference at one means equality of the two sample means, not equality of their full distributions. The original plot's \emph{Natural Baseline} label denotes this reference. Ratios both above and below one argue against a universal low-entropy detection rule.}
\label{fig:gcm_ratio}
\end{figure}

Figure~\ref{fig:gcm_ratio} reports the ratio of mean image-level GCM entropy in real and synthetic subsets. The direction of the difference depends on the source: the displayed ratios are approximately 1.35 for DALL-E~2 and GLIDE, but 0.92 for Wukong and below one for several other subsets. These observations are inconsistent with a universal rule that generated images have lower local entropy. Instead, they suggest that a simple local statistic can shift in either direction across generation and data-processing pipelines.

\begin{figure}[p]
\centering
\includegraphics[width=0.98\linewidth,height=0.83\textheight,keepaspectratio]{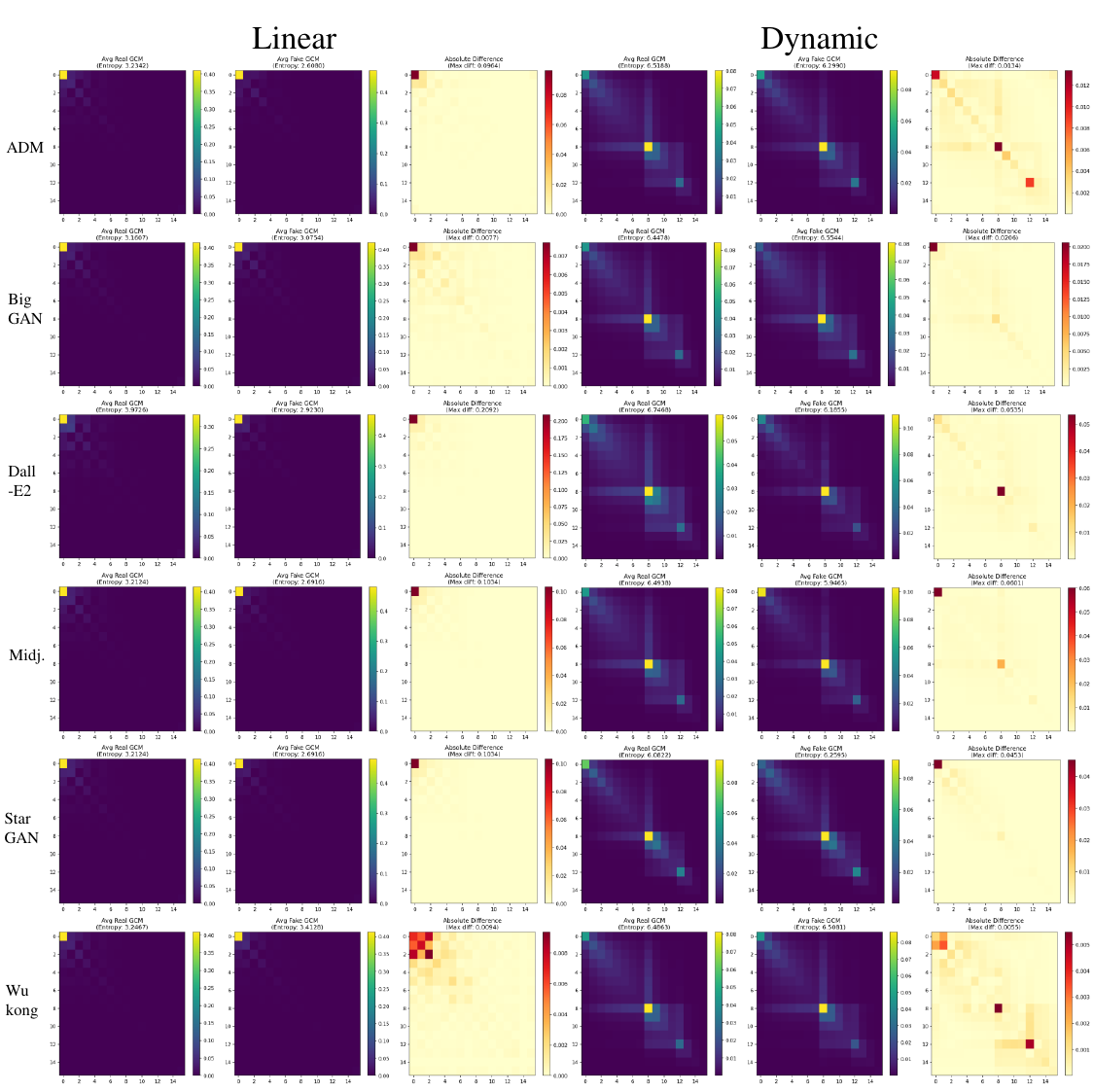}
\caption{GCM visualizations under linear-range quantization (left) and dynamic stratification (right). Each group shows the real-image mean matrix, synthetic-image mean matrix, and absolute difference. Color ranges vary between panels. The structures reflect both image statistics and quantization; their appearance is not evidence that the neural detector uses the same features.}
\label{fig:gcm_comparison_grid}
\end{figure}

Figure~\ref{fig:gcm_comparison_grid} compares the linear and stratified visualizations. Linear quantization concentrates much of the mass near the low-gradient corner, whereas stratification makes other transitions easier to inspect. The resulting differences depend on the quantizer: for example, the same subset need not exhibit the same ordering of mean entropies under both representations. Figure~\ref{fig:gcm_stats} further shows substantial overlap between real and generated entropy distributions. A scalar entropy threshold is therefore not a reliable universal detector in these examples.

These figures were produced during exploratory analysis, including revisions to range clipping and quantization; the saved linear implementation additionally uses 99th-percentile clipping. They should be read as related diagnostics, not as a matched comparison of quantizers on an identical fixed sample. Content, image resizing, compression, and the composition of real-image subsets can all contribute to the differences. No statistical significance claim or attribution to a unique generation mechanism follows from these plots. Their useful implication is narrower: local adjacency statistics are worth investigating, while the learned detector's actual cues require separate analysis.

\begin{landscape}
\thispagestyle{plain}
\begin{figure}[p]
\centering
\includegraphics[width=0.98\linewidth]{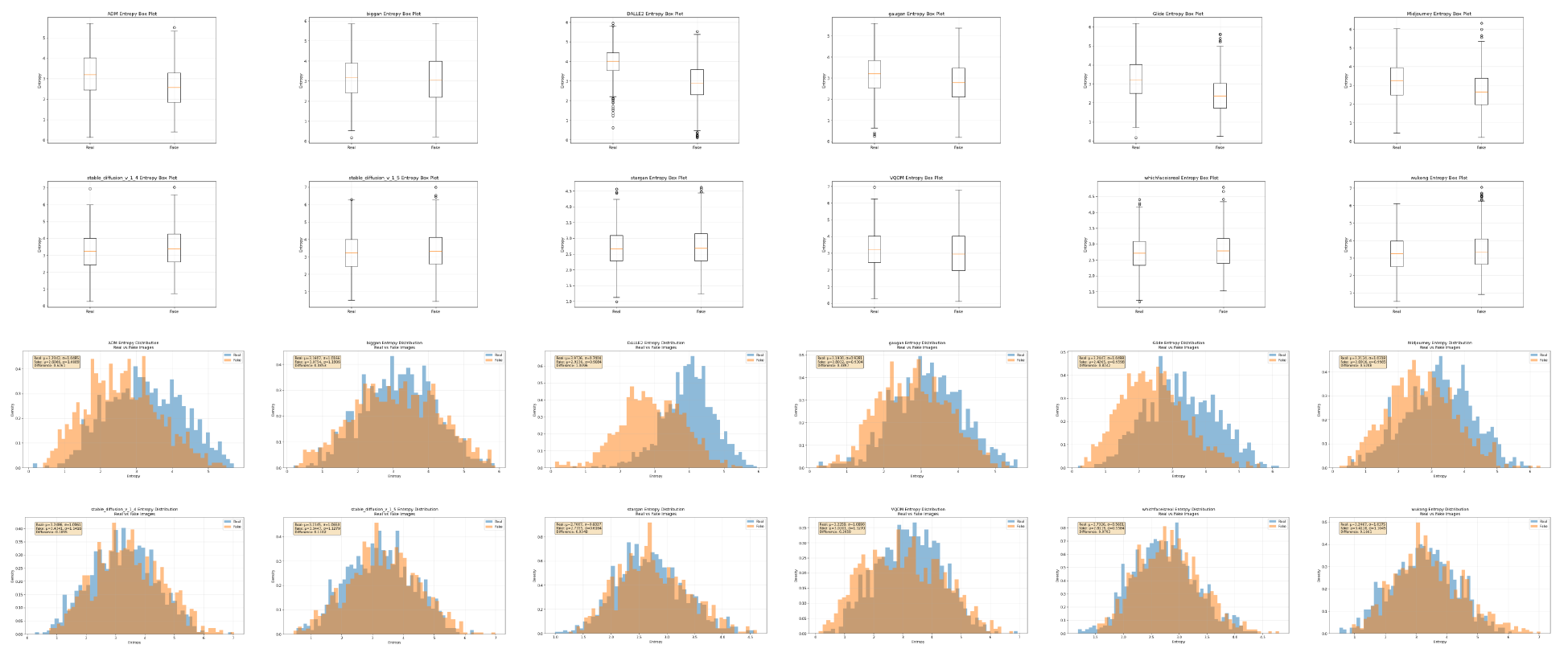}
\caption{Exploratory image-level GCM entropy distributions across the examined subsets: box plots in the upper two rows and histograms in the lower two rows. Real and synthetic distributions overlap, and their relative positions vary by subset. These plots complement the ratios in Figure~\ref{fig:gcm_ratio}; they are not a detection benchmark or a statistical test of the learned representation.}
\label{fig:gcm_stats}
\end{figure}
\end{landscape}

\clearpage
\section{Patch Scores and Failure Cases}
\label{sec:appendix_vis}

The following examples use a ResNet-50 patch detector trained on ProGAN. Each map places synthetic-class scores back on the image. They visualize predictions, not pixel-level forgery annotations.

\paragraph{Spatially heterogeneous evidence.}
In Figure~\ref{fig:inference_vis}, real photographs receive lower image-mean scores than the generated examples, but patches within one image need not agree. The Midjourney example contains mixed local responses while its average remains above the threshold. Aggregation combines these responses without establishing that every high-score region contains an identifiable artifact.

\begin{center}
\includegraphics[width=0.94\linewidth,height=0.70\textheight,keepaspectratio]{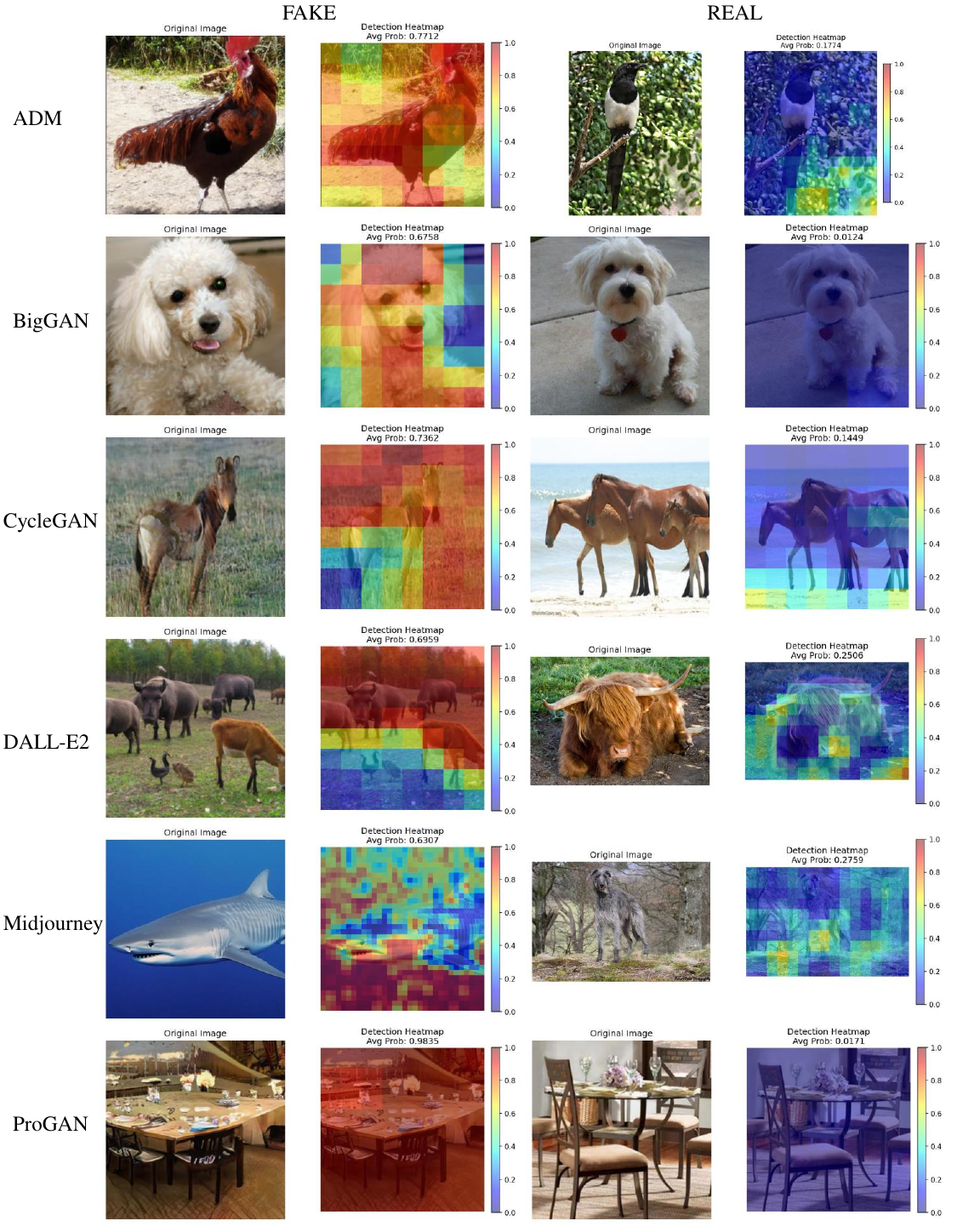}
\captionof{figure}{Successful image-level predictions with patch-wise score maps. The left group contains generated images and the right group real photographs; warmer colors indicate higher synthetic-class scores. The shown average is the image-level score. Local colors are model predictions, not verified artifact locations.}
\label{fig:inference_vis}
\end{center}

\clearpage
\paragraph{Failures and resizing sensitivity.}
The Midjourney examples in Figure~\ref{fig:failure_cases} are false negatives (0.2954 and 0.4649); the real GauGAN-subset examples are false positives (0.8454 and 0.9415). For the two real-image pairs below, resizing to $256\times256$ changes the scores from 0.1034 to 0.5818 and from 0.4123 to 0.9216, crossing the 0.5 threshold. These examples show processing sensitivity, but do not isolate interpolation, geometric distortion, or changed crop coverage as the cause.

\begin{center}
\includegraphics[width=0.94\linewidth,height=0.73\textheight,keepaspectratio]{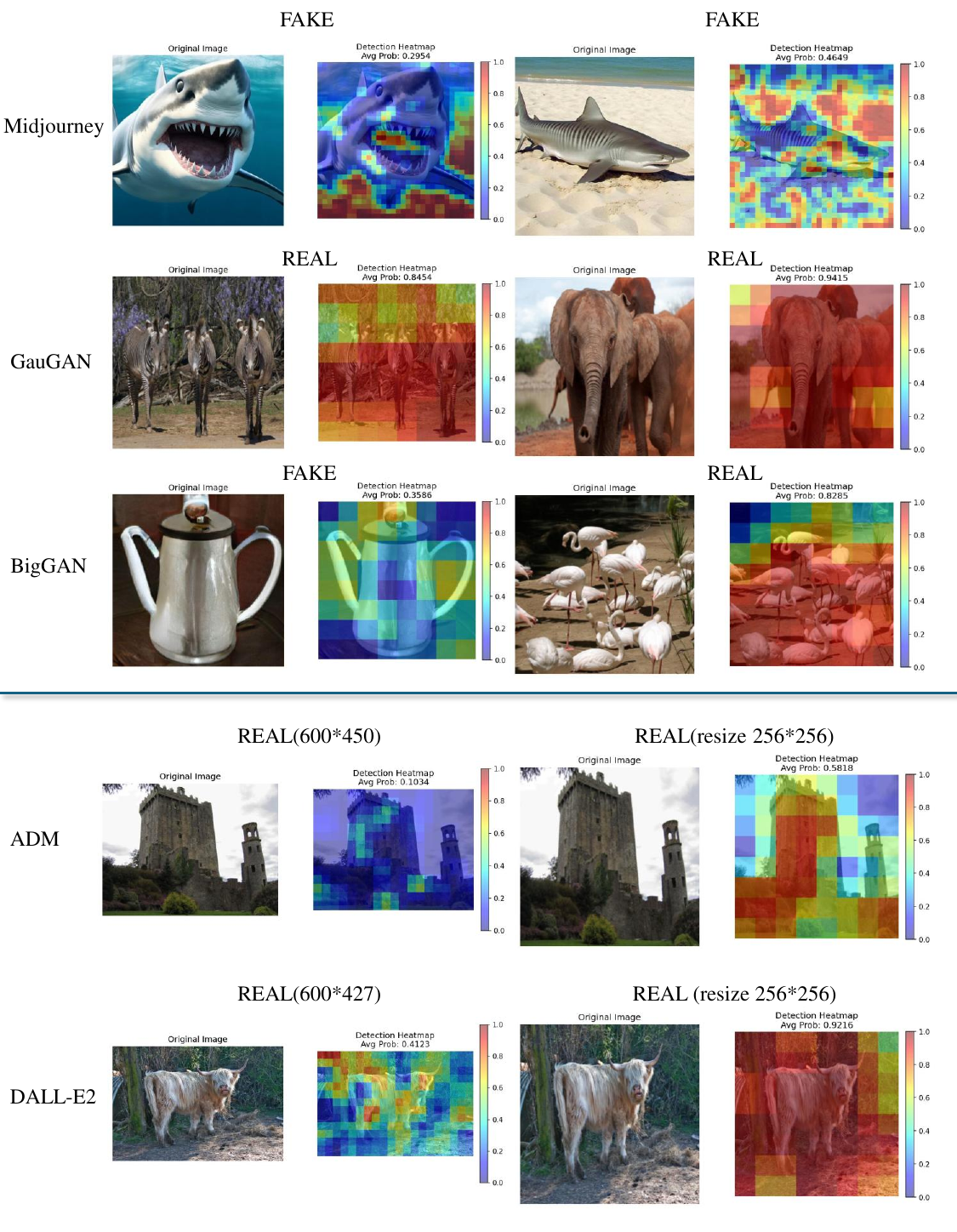}
\captionof{figure}{Selected false negatives, false positives, and resizing-sensitive predictions. REAL/FAKE denotes the image label, not the predicted class. The upper examples show errors across three subsets; the lower pairs are real images before and after resizing. These are illustrative cases rather than an estimate of failure frequencies.}
\label{fig:failure_cases}
\end{center}